%% file: icml.tex
\documentclass{article}

\usepackage{microtype}
\usepackage{graphicx}
\usepackage{subcaption}
\usepackage{booktabs} 

\usepackage{hyperref}

\usepackage[preprint]{icml2026}

\usepackage{amsmath}
\usepackage{amssymb}
\usepackage{mathtools}
\usepackage{amsthm}
\usepackage{graphicx}
\usepackage{wrapfig}
\usepackage[figuresright]{rotating}
\usepackage{makecell}
\usepackage{diagbox}   
\usepackage{multirow}
\usepackage[T1]{fontenc}
\usepackage{listings}
\usepackage{xcolor}

\usepackage[capitalize,noabbrev]{cleveref}

\theoremstyle{plain}
\newtheorem{theorem}{Theorem}[section]
\newtheorem{proposition}[theorem]{Proposition}
\newtheorem{lemma}[theorem]{Lemma}
\newtheorem{corollary}[theorem]{Corollary}
\theoremstyle{definition}
\newtheorem{definition}[theorem]{Definition}
\newtheorem{assumption}[theorem]{Assumption}
\theoremstyle{remark}
\newtheorem{remark}[theorem]{Remark}

\usepackage[textsize=tiny]{todonotes}

\icmltitlerunning{Beyond Scores: Diagnostic LLM Evaluation via Fine-Grained Abilities}

\begin{document}

\twocolumn[
  \icmltitle{Beyond Scores: Diagnostic LLM Evaluation \\ via Fine-Grained Abilities}



  \icmlsetsymbol{equal}{*}

  \begin{icmlauthorlist}
    \icmlauthor{Xu Zhang}{1,2}
    \icmlauthor{Xudong Gong}{1,2}
    \icmlauthor{Jiacheng Qin}{1,2}
    \icmlauthor{Qiang Wang}{1,2}
    \icmlauthor{JiaQi Liao}{1,2}
    \icmlauthor{Zhe Wang}{4}
    \icmlauthor{Dawei Feng}{1,2}
    \icmlauthor{Bo Ding}{1,3}
  \end{icmlauthorlist}

  \icmlaffiliation{1}{College of Computer Science and Technology, National University of Defense Technology, Changsha, Hunan, China}
  \icmlaffiliation{2}{State Key Laboratory of Complex \& Critical Software Environment, Changsha, Hunan, China}
  \icmlaffiliation{3}{National Key Laboratory of Parallel and Distributed Computing, Changsha, Hunan, China}
  \icmlaffiliation{4}{School of Humanities and Social Sciences, School of Public Administration, Beihang University, Beijing, China}

  \icmlcorrespondingauthor{Dawei Feng}{davyfeng.c@qq.com}

  \icmlkeywords{Machine Learning, ICML}

  \vskip 0.3in
]



\printAffiliationsAndNotice{}  

\input{sections/abstract.tex}

\input{sections/intro.tex}

\input{sections/method.tex}

\input{sections/result.tex}

\input{sections/relate.tex}

\input{sections/conclusion.tex}

\section*{Impact Statements}
This paper presents work whose goal is to advance the field of machine learning. There are many potential societal consequences of our work, none of which we feel must be specifically highlighted here.

\bibliography{bib.bib}
\bibliographystyle{icml2026}

\newpage
\appendix
\onecolumn
\input{appendix/prompt.tex}
\input{appendix/taxonomy.tex}
\input{appendix/domains.tex}
\input{appendix/setup.tex}
\input{appendix/prediction.tex}

\end{document}

%% file: sections/abstract.tex
\begin{abstract}

Current evaluations of large language models aggregate performance across diverse tasks into single scores. This obscures fine-grained ability variation, limiting targeted model improvement and ability-guided selection for specific tasks. Motivated by this gap, we propose a cognitive diagnostic framework that estimates model abilities across multiple fine-grained dimensions. For mathematics, we construct a 35-dimensional ability taxonomy grounded in cognitive theory and domain knowledge. The framework employs multidimensional Item Response Theory with an item-ability association matrix to estimate fine-grained ability levels, which in turn enable prediction of performance on unseen items (questions of benchmark). Evaluated on 41 models, our approach demonstrates strong criterion validity, consistent ability estimates across benchmarks, and accurate prediction of unseen items with AUC ranging from 0.80 to 0.89 within benchmarks and from 0.77 to 0.86 across benchmarks, substantially exceeding trivial baselines. The framework generalizes across scientific domains, producing consistent diagnostic performance in physics (27 dimensions), chemistry (58 dimensions), and computer science (12 dimensions). This work establishes a principled framework for fine-grained assessment of abilities, with potential applications in targeted training, ability-guided model selection, and ability-aware benchmark design.

\end{abstract}

%% file: sections/intro.tex
\section{Introduction}
\label{sec:intro}

\begin{figure*}[htbp]
    \centering
    \resizebox{0.9\textwidth}{!}{
        \includegraphics{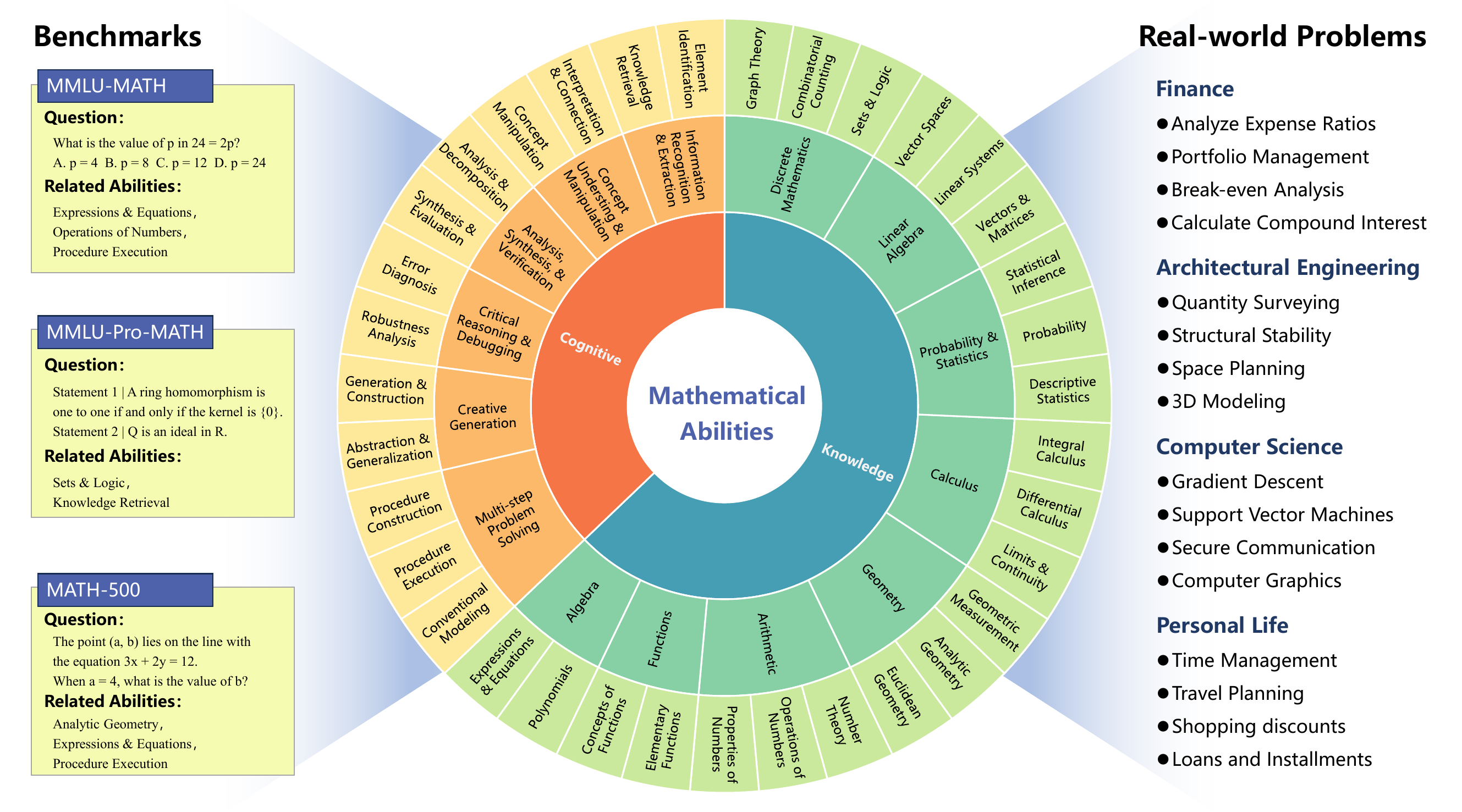}
    }
    \caption{A diagnostic framework for evaluating LLMs via fine-grained mathematical abilities. Centered on Bloom's taxonomy, we decompose Mathematical Abilities into two core dimensions: Knowledge (e.g., Algebra, Calculus) and Cognitive (e.g., Multi-step Problem Solving, Critical Reasoning). Each is further subdivided into interpretable sub-skills. The framework links benchmark questions (left) to their underlying abilities and maps them to real-world problem domains (right), enabling granular diagnosis beyond aggregate scores and bridging the gap between evaluation benchmarks and practical task demands.}
    \label{fig:finegrain}
\end{figure*}

Large language models (LLMs) have demonstrated remarkable progress across diverse domains, with performance on standardized benchmarks approaching or surpassing human levels in specific tasks \cite{qin2025soapfl, wu2024boosting, wang2025Large}. However, this progress is predominantly quantified through aggregate scores that condense performance across diverse tasks into single numerical summaries. Because a model's task-level capabilities emerge from the composition of fine-grained abilities, such as recalling algebraic identities (domain knowledge) or planning multi-step derivations (cognitive processes), these aggregate metrics obscure fine-grained ability variation and limit both targeted model improvement and ability-guided selection for specific tasks. Prior studies reinforce this view: Griot et al. \cite{griot2025large} highlight missing metacognitive skills in medical reasoning; Cao et al. \cite{cao2025toward} advocate shifting from task-based to ability-centric evaluation; and Liu et al. \cite{liu2024finemath} demonstrate that aggregate accuracy masks mastery of specific concepts.

Cognitive diagnostic methods from educational assessment, including Classical Test Theory (CTT) \cite{devellis2006classical}, Item Response Theory (IRT) \cite{harvey1999item}, and multidimensional IRT (mIRT) \cite{ackerman2014multidimensional}, were originally developed to diagnose students' fine-grained abilities from their responses to assessment items. Recent work has successfully adapted these methods to LLM evaluation \cite{liu2023what,yao2025je,zhao2025can,zhou2025lost}, establishing IRT as a viable framework for model assessment. However, these adaptations treat benchmark items as atomic units and lack explicit mappings between items and the fine-grained abilities they engage. Consequently, they remain confined to coarse-grained performance summaries and cannot support ability-level diagnosis for targeted improvement or task-aligned selection.

To overcome this limitation, we propose a cognitive diagnostic framework that estimates model abilities across multiple fine-grained dimensions. The framework constructs domain-specific ability taxonomies grounded in cognitive theory and domain knowledge; for mathematics, this yields a 35-dimensional taxonomy. An item-ability association matrix (Q-matrix) encodes which fine-grained abilities each benchmark item (question) engages. The framework employs mIRT with the Q-matrix to estimate fine-grained ability levels. These estimates enable both ability-aware model selection aligned with task requirements and accurate prediction of performance on unseen items.

Evaluated on 41 models, the framework demonstrates strong criterion validity, consistent ability estimates across benchmarks, and accurate prediction of unseen items with AUC ranging from 0.80 to 0.89 within benchmarks and from 0.77 to 0.86 across benchmarks. The framework generalizes across scientific domains, producing consistent diagnostic performance in physics (27 dimensions), chemistry (58 dimensions), and computer science (12 dimensions). 
Our work establishes a principled diagnostic framework that produces interpretable fine-grained ability profiles for LLMs, with potential to inform targeted training, ability-guided model selection, ability-aware benchmark design, and alignment between benchmark evaluations and real-world task demands (Figure~\ref{fig:finegrain}).

%% file: sections/method.tex
\section{Diagnostic LLM Evaluation via Fine-Grained Abilities}
\label{sec:method}

\subsection{Overview}
\label{sec:framework}

\begin{figure}[htbp]
    \centering
    \resizebox{\linewidth}{!}{
        \includegraphics{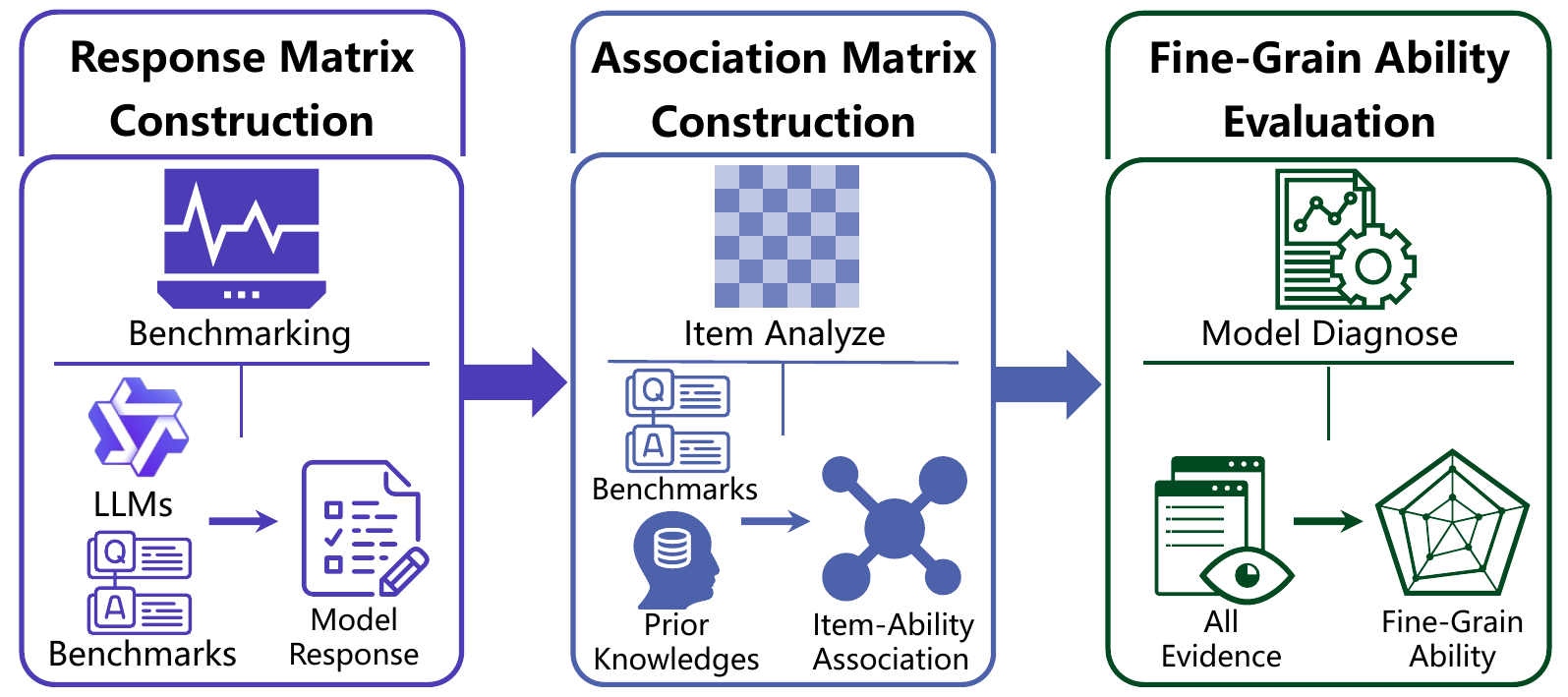}
    }
    \caption{Architecture of the fine-grained ability assessment framework. The framework comprises three stages: (a) Response Matrix Construction, where models are evaluated on benchmarks to generate a binary model-item response matrix; (b) Association Matrix Construction, in which each item of benchmark is mapped to fine-grained abilities (e.g., domain knowledge or cognitive processes) via expert annotation and prior knowledge, forming a Q-matrix; and (c) Fine-Grain Ability Evaluation, where a cognitive diagnostic model estimates latent ability profiles by jointly analyzing the response and association matrices.}
    \label{fig:framework}
\end{figure}

We describe a diagnostic evaluation framework that decomposes model performance on existing benchmarks into fine-grained abilities. These abilities include domain knowledge (e.g., algebraic rules, geometric facts) and cognitive processes (e.g., multi-step reasoning, problem decomposition). As shown in Figure~\ref{fig:framework}, the framework comprises three components.
(1) A model-item response matrix (R-matrix) derived from existing benchmark evaluations without modifying the original assessment protocol, where entries indicate model correctness on individual items;
(2) An item-ability association matrix (Q-matrix) that explicitly maps each benchmark item to the fine-grained abilities it engages;
(3) Multidimensional Item Response Theory (mIRT) that estimates latent ability levels for each model based on the Q-matrix and R-matrix.

\textbf{Model-Item Response Matrix Construction.}
We evaluate each target model on all items in a given benchmark and record its responses in binary form, where $1$ indicates a correct answer and $0$ indicates an incorrect or incomplete response. This yields an $m \times n$ response matrix $\mathbf{R}$, with $m$ denoting the number of models and $n$ the number of benchmark items. This matrix serves as the empirical basis for subsequent diagnostic inference.

\textbf{Item-Ability Association Matrix Construction.}
The Q-matrix, denoted $\mathbf{Q}\in \{0, 1\}^{n\times k}$, specifies which fine-grained abilities are required to correctly solve each item of benchmarks. Here, $k$ is the total number of predefined abilities, for example, "recalling trigonometric identities" or "constructing a proof by contradiction." An entry $q_{ij}=1$ indicates that item $i$ depends on ability $j$; otherwise, $q_{ij}=0$. The Q-matrix is constructed through a combination of expert annotation and LLM-assisted ability labeling, guided by our taxonomy of mathematical competence (Section \ref{sec:q}).

\textbf{Fine-Grained Ability Estimation.}
Given the response matrix $\mathbf{R}$ and the Q-matrix $\mathbf{Q}$, we estimate each model's latent ability profile (a $k$-dimensional vector quantifying its proficiency across fine-grained abilities) using a multidimensional cognitive diagnostic model grounded in mIRT (Section \ref{sec:model}). Specifically, the probability that model $m$ answers item $i$ correctly is modeled as a function of its proficiency across the $k$ abilities, along with the item's discrimination and difficulty parameters. These profiles enable interpretable diagnosis beyond aggregate scores.

\subsection{Fine-Grained Ability Taxonomy}
\label{sec:taxonomy}

As illustrated in Figure~\ref{fig:finegrain}, our ability taxonomy is inspired by Bloom's Taxonomy \cite{forehand2010bloom} and organizes model competence along two dimension categories: domain knowledge and cognitive processes. Each dimension is further decomposed into fine-grained, interpretable subcategories that serve as the basis for diagnostic evaluation.

\textbf{Domain knowledge} refers to the model's mastery of static, factual, or procedural content within a specific discipline. In the context of mathematical reasoning, we define knowledge subcategories, such as Algebra, Functions, Arithmetic, and Geometry, by aligning with standard mathematical curricula and the scope of the benchmarks used. 

\textbf{Cognitive processes} capture the model's capacity to apply knowledge in task-solving contexts. Rather than directly adopting Bloom's original cognitive levels, which were designed for human learners, we adapt them to the LLM setting by focusing on observable reasoning behaviors. This yields six operationalized cognitive dimensions:
(1) Information Recognition and Extraction,
(2) Concept Understanding and Manipulation,
(3) Analysis, Synthesis, and Verification,
(4) Critical Reasoning and Debugging,
(5) Creative Generation, and
(6) Multi-step Problem Solving.
These dimensions are defined based on the types of reasoning required to solve benchmark items, ensuring alignment between ability definitions and empirical evaluation.

Each benchmark item is linked to one or more abilities spanning both dimensions, with the full annotation protocol described in Section \ref{sec:q}.

\subsection{Constructing the Q-Matrix via Expert-Supervised LLM Annotation}
\label{sec:q}

To enable diagnostic evaluation on existing benchmarks, we construct the Q-matrix, an item-ability association matrix that maps each benchmark item to fine-grained abilities along both domain knowledge and cognitive processes dimensions (Section \ref{sec:taxonomy}), using a human-in-the-loop annotation pipeline. This pipeline leverages LLMs for scalable ability labeling while incorporating domain experts at critical stages to ensure conceptual fidelity and alignment with educational standards. The process proceeds in three stages.

\textbf{(a) Ability Pool Generation.}
For each benchmark item, we prompt an LLM to identify the fine-grained abilities it assesses, covering both domain knowledge (e.g., Algebra, Geometry) and cognitive processes (e.g., Multi-step Problem Solving, Logical Inference). The union of all identified abilities across items forms an initial ability pool.

\textbf{(b) Hierarchical Ability System Refinement.}
Guided by standard mathematical curricula and our adapted Bloom's Taxonomy, we design structured prompts to organize the ability pool into a hierarchical system spanning multiple granularities (e.g., coarse: "Algebra"; fine: "polynomials"). This system is then fully reviewed and revised by domain experts to ensure conceptual coherence and alignment with educational standards.

\textbf{(c) Q-Matrix Construction.}
From the refined hierarchy, we select a target granularity level as the final set of evaluation dimensions. An LLM is then prompted to map each benchmark item to the relevant dimensions in this set, producing a preliminary Q-matrix.

Throughout the process, stage (b) undergoes full expert review, while stages (a) and (c) are spot-checked by experts at a 10\% random sampling rate, with corrections applied to the entire dataset when systematic errors are detected. All prompts used in this workflow are provided in Appendix \ref{app:prompt}.

\subsection{Estimating Fine-Grained Abilities via mIRT}
\label{sec:model}

Latent trait measurement is a well-established paradigm in psychometrics, with mature methodologies in educational assessment and cognitive diagnosis. To estimate fine-grained ability profiles of LLMs, we adapt mIRT to model the relationship between model responses and underlying competencies spanning both domain knowledge and cognitive processes.

Item Response Theory (IRT) posits a probabilistic relationship between an examinee's latent traits and their observed binary responses. Our framework enables two key functionalities:
(1) estimating continuous latent ability profiles from model-item response data, and
(2) predicting performance on unseen items by jointly modeling ability parameters and item characteristics, such as discrimination and difficulty.

\begin{figure}[htbp]
    \centering
    \includegraphics[width=0.8\linewidth]{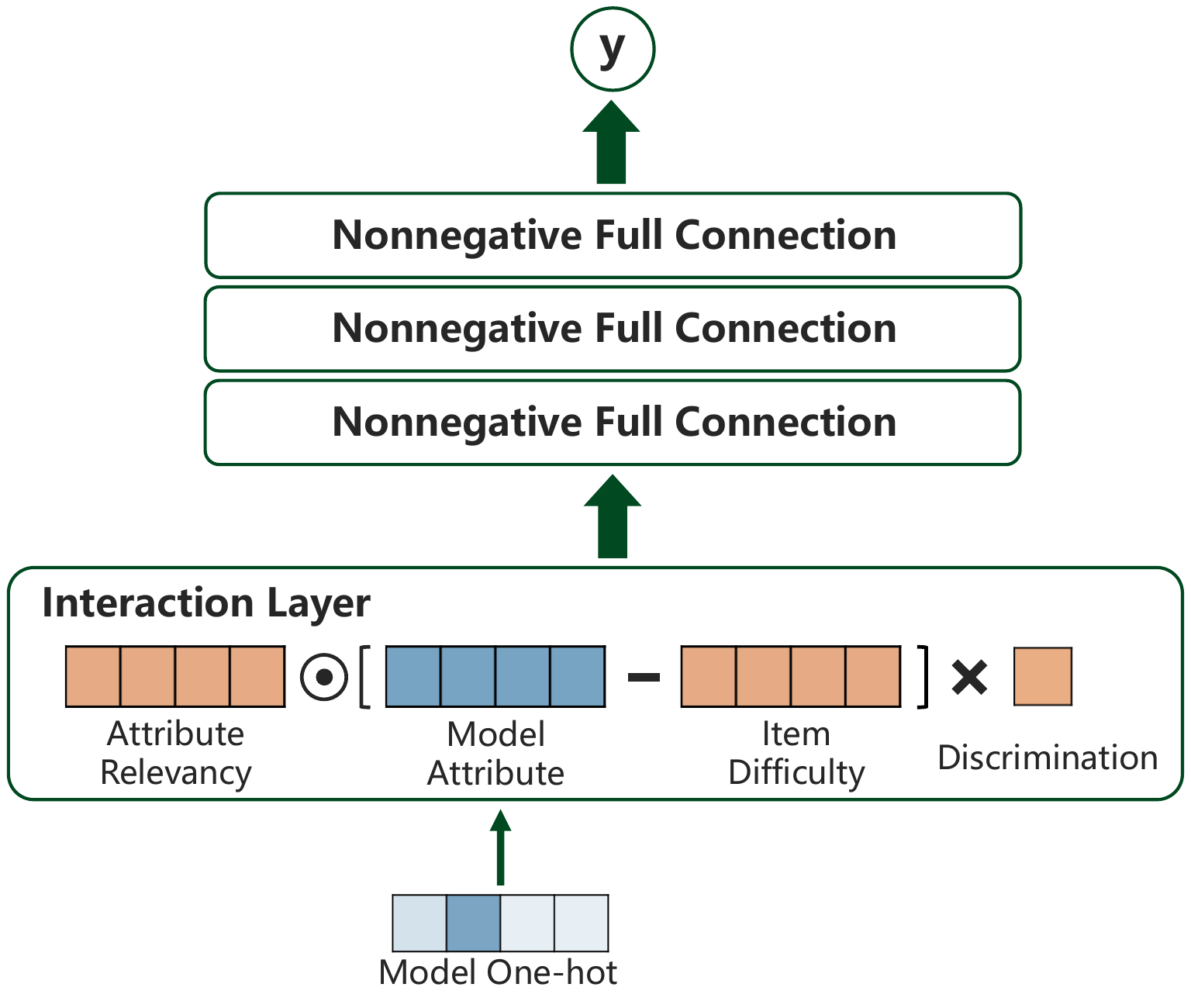}
    \caption{Structure of the IRT-based fine-grained ability evaluation model. Based on the IRT method, a simple fully connected neural network is implemented on the Interaction Layer to fit model performance in downstream tasks, thereby achieving fine-grained ability evaluation.}
    \label{fig:irt}
\end{figure}

We adapt the NeuralCD framework~\cite{wang2023neuralcd}, a neural extension of mIRT, to estimate LLM ability profiles based on our fine-grained ability taxonomy (Section~\ref{sec:taxonomy}). As shown in Figure~\ref{fig:irt}, the model enforces the monotonicity assumption of IRT, which states that higher ability levels should yield higher probabilities of correct responses, by constraining all weights in the three fully connected layers to be non-negative. This ensures that the predicted response probability is a monotonically increasing function of each latent ability dimension.
Then, the interaction layer is inspired by mIRT and formalized as follows:

\begin{equation}
\label{eq:intersection}
    \boldsymbol{\mathit{x}}= \boldsymbol{q} \odot (\boldsymbol{a}-\boldsymbol{I}_{\mathrm{diff}})\times \boldsymbol{I}_{\mathrm{disc}}
\end{equation}

Here, $\boldsymbol{q}\in \{0,1\}^K$ is the item-ability association vector (i.e., a row of the Q-matrix), 
$\boldsymbol{\mathit{a}}\in [0,1]^K$ denotes the model's latent ability profile, 
$\boldsymbol{I}_{\mathrm{diff}}\in[0,1]^{K}$ represents item-specific difficulty parameters, 
$I_{\mathrm{disc}}\in[0,1]^{K}$ is the item discrimination scalar,
$\odot$ denotes element-wise multiplication, 
and $\boldsymbol{\mathit{x}}\in\mathbb{R}^K$ is the input to the subsequent MLP layer.

The model's ability profile $\boldsymbol{\mathit{a}}$ is then computed via:

\begin{equation}
\label{eq:irtatom}
    \boldsymbol{\mathit{a}}= \mathrm{sigmoid}(\boldsymbol{\mathit{x}}_m \times \mathbf{A})
\end{equation}

where $\boldsymbol{\mathit{x}}_m\in\{0,1\}^{1\times M}$ is a one-hot vector indicating the $m$-th model, and $\mathbf{A}\in\mathbb{R}^{M\times K}$ is a trainable matrix whose rows represent the fine-grained ability profiles of all $M$ evaluated models.

This formulation allows us to jointly learn item parameters $(\boldsymbol{I_{\mathrm{diff}}}, \boldsymbol{I_{\mathrm{disc}}})$ and model ability profiles $\mathbf{A}$ from observed response data, yielding interpretable, multi-dimensional diagnostics beyond aggregate scores.

%% file: sections/result.tex
\section{Experiment and Result Analysis}
\label{sec:result}

\subsection{Experiment Setup}
\label{sec:setup}

\begin{figure*}[htbp]
    \centering
    \begin{subfigure}{0.48\textwidth}
        \centering
        \includegraphics[width=\linewidth]{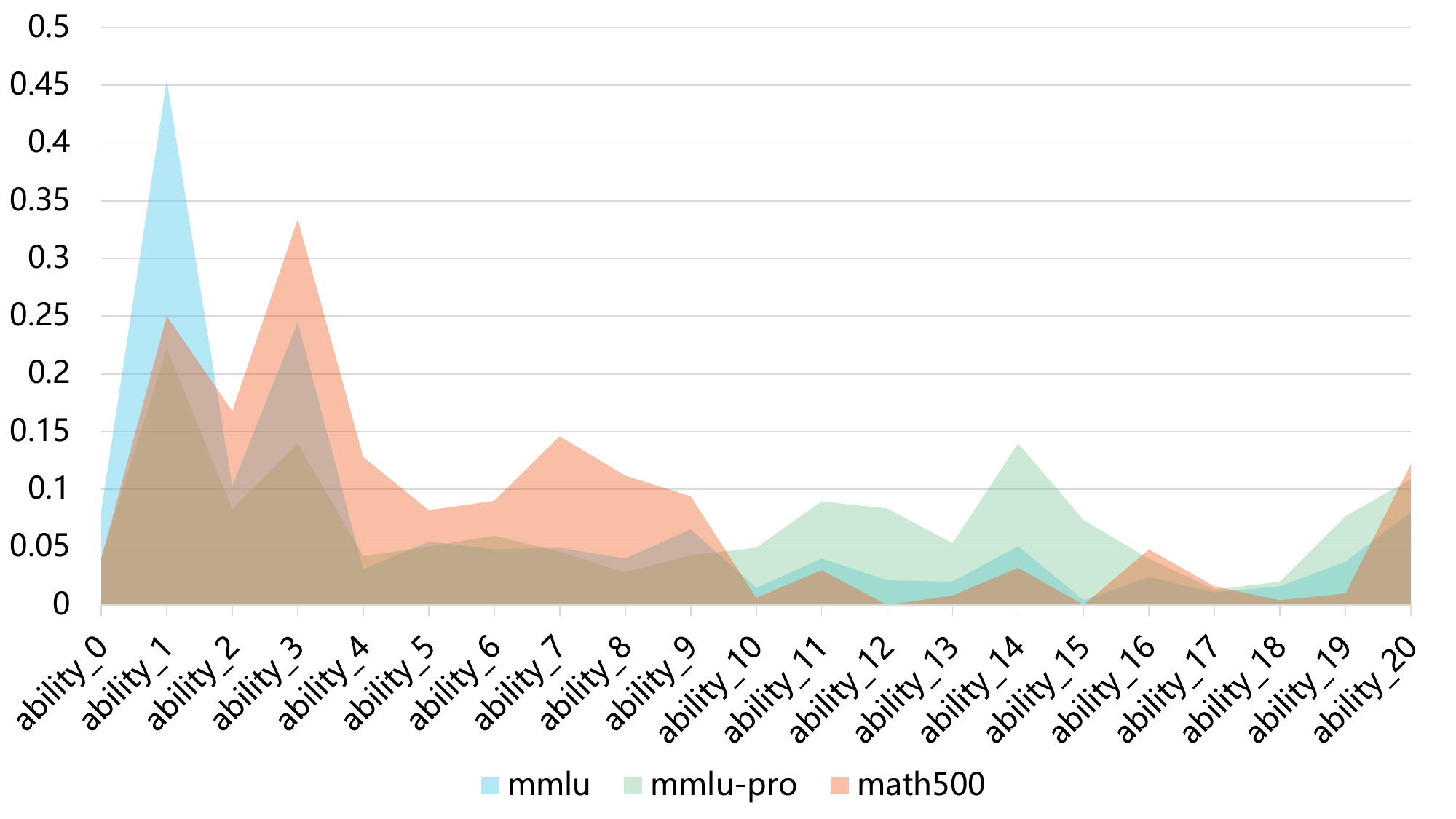}
        \caption{Distributions of Knowledge Abilities}
        \label{fig:knowledge}
    \end{subfigure}
    \begin{subfigure}{0.48\textwidth}
        \centering
        \includegraphics[width=\linewidth]{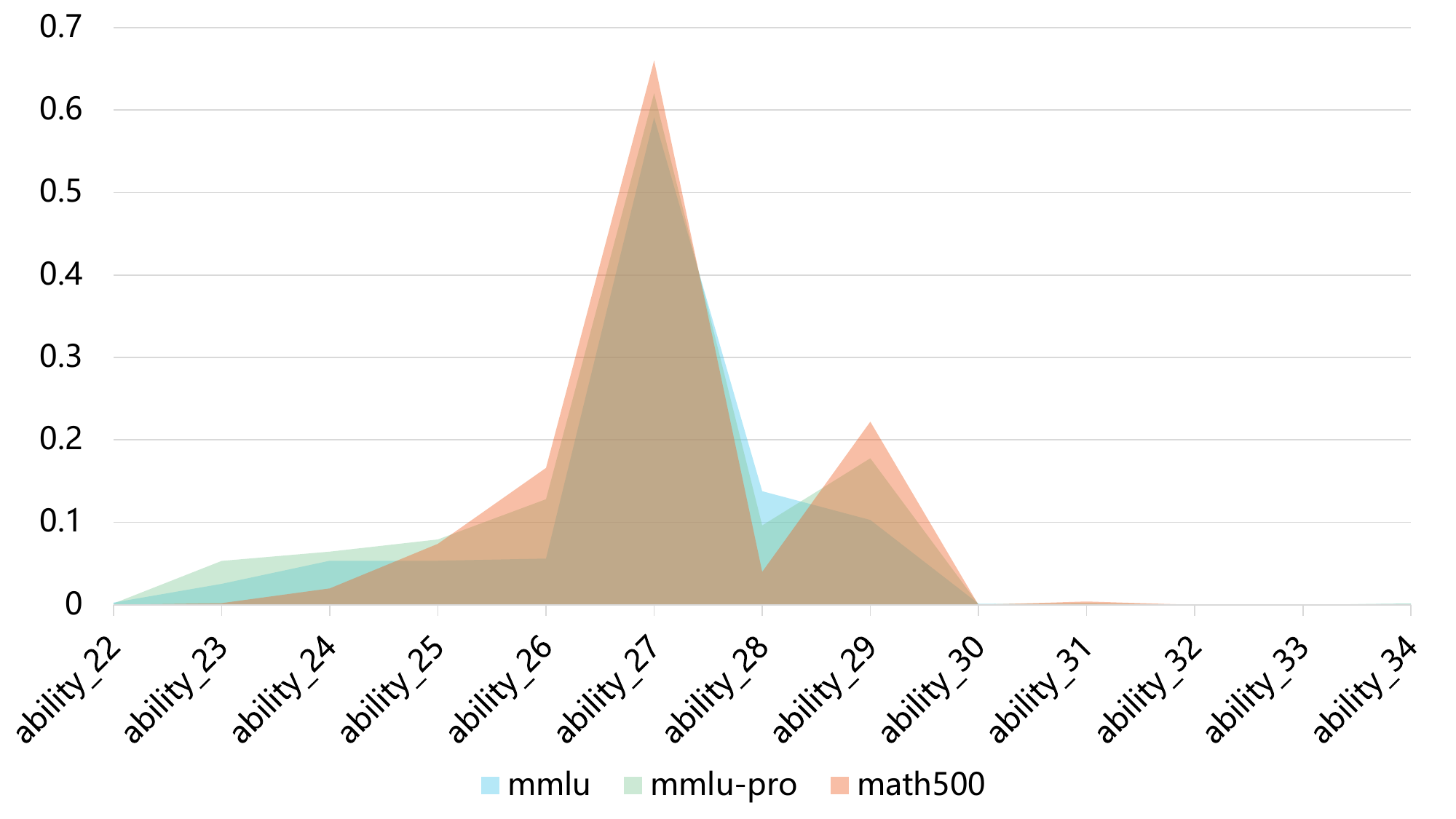}
        \caption{Distributions of Cognitive Abilities}
        \label{fig:cognitive}
    \end{subfigure}
    \caption{\textbf{Distribution of item coverage across 35 fine-grained abilities in three mathematical benchmarks.} The distributions are inconsistent across knowledge-oriented and cognition-oriented dimensions, with pronounced differences among individual dimensions. { \bf Left}: benchmark-knowledge ability associations, indicating distinct assessment emphases across benchmarks. {\bf Right}: benchmark-cognition ability associations, exhibiting greater overlap in assessment tendencies while retaining benchmark-specific emphases.}
    \label{fig:distribution}
\end{figure*}

We evaluate our fine-grained diagnostic framework on a diverse set of benchmarks and models. The effectiveness of the framework is validated from three perspectives: (1) diagnostic utility of the fine-grained ability estimates, (2) cross-benchmark consistency of ability estimates, and (3) predictive performance on held-out items. We further assess cross-domain generalization by extending the evaluation to three non-mathematical subjects.

For the mathematical domain, we use three widely adopted benchmarks: MMLU-MATH, MMLU-Pro-MATH, and MATH500.
Our evaluation encompasses 35 open-source LLMs and 6 proprietary online LLMs. 
All models are evaluated through the OpenCompass benchmarking suite, which provides consistent prompt templates, decoding strategies, and scoring rules across benchmarks.
For each model, a binary response matrix is generated by evaluating it on all benchmark items, with correct answers encoded as 1 and incorrect responses as 0.

As shown in Figure~\ref{fig:distribution}, the three mathematical benchmarks exhibit distinct distributions of item coverage across ability dimensions. Detailed coverage statistics are provided in Appendix~\ref{app:coverage}. This heterogeneity motivates our cross-benchmark analysis, which evaluates whether ability profiles generalize across these differing distributions.

For ability estimation, we implement the mIRT-based evaluation model described in Section \ref{sec:model}, a three-layer fully connected network with hidden dimensions (256, 1024, 128). The model employs CrossEntropyLoss as the loss function and is optimized using Adam with an initial learning rate of 0.001 (decayed by 0.8) and a batch size of 256. All experiments use an NVIDIA RTX 3090 GPU.

Further details of the experimental setup are provided in Appendix~\ref{app:setup}.

\subsection{Validity of the Fine-Grained Diagnostic Framework}
\label{sec:reliablity}

Validity refers to the extent to which a measurement tool accurately reflects the construct it is intended to assess. We focus on criterion-related validity, which evaluates whether estimated ability scores correlate with external measures of performance on relevant tasks. To this end, we compute Spearman's rank correlation coefficient ($\rho$) between the estimated ability score for each dimension and the model's accuracy on the set of benchmark items linked to that dimension.
For each dimension, accuracy is defined as the mean binary accuracy across all items associated with that dimension in the Q-matrix. High correlation indicates that the estimated ability levels reliably capture the model's actual performance on relevant evaluation items.
The detailed computation procedure is provided in Appendix \ref{app:corr}. 

Results across the three mathematical benchmarks, MMLU-MATH, MMLU-Pro-MATH, and MATH500, are visualized in Figure~\ref{fig:downstream}, with full numerical results in Appendix \ref{app:taxonomy}. We define correlation strength as follows: strong ($\rho > 0.7$), moderate ($0.5 < \rho \le 0.7$), weak ($0.3 < \rho \le 0.5$), and non-significant ($\rho \le 0.3$ or $p > 0.05$).

To assess the robustness of our diagnostic framework, we examine the consistency of criterion validity for each of the 35 ability dimensions across all three mathematical benchmarks. This analysis reveals two key patterns.

First, 18 dimensions exhibit strong criterion correlation in at least two benchmarks, demonstrating stable diagnostic power. Notably, 9 dimensions maintain strong correlation across all three benchmarks. The consistency of these signals across diverse item sets confirms that our framework captures generalizable model competencies.

\begin{figure*}[htbp]
    \centering
    \resizebox{\linewidth}{!}{
        \includegraphics{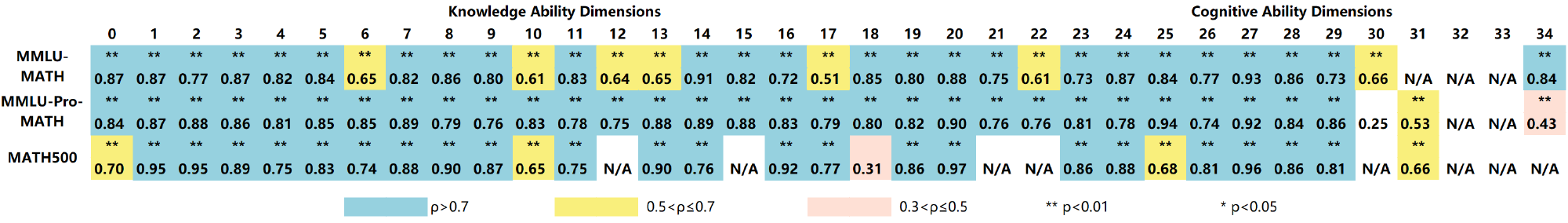}
    }
    \caption{
        \textbf{Correlation between estimated fine-grained abilities and model performance on associated benchmark items.}
        Spearman's rank correlation coefficient ($\rho$) measures the association between each ability dimension (columns) and model accuracy on its linked items across three benchmarks (rows).
        \textbf{Color coding}: Blue indicates $\rho > 0.7$ (strong), yellow indicates $0.5 < \rho \le 0.7$ (moderate), and red indicates $0.3 < \rho \le 0.5$ (weak).
        \textbf{Significance markers}: ** denotes $p<0.01$, * denotes $p<0.05$.
        \textbf{N/A} indicates that no items in the benchmark are associated with that dimension (i.e., invalid for evaluation).
        Dimensions with consistent strong correlation across benchmarks reflect stable diagnostic signals; weak correlations often arise from low item coverage (<10 items). 
    }
    \label{fig:downstream}
\end{figure*}

Second, dimensions with weak or non-significant correlations are consistently associated with sparse item coverage. As shown in Figure~\ref{fig:distribution}, these dimensions correspond to subsets containing fewer than 10 questions, which is less than 5\% of the total item pool. For such low-sample dimensions, correlation estimates are inherently unstable, leading to reduced validity. Crucially, no dimension shows high correlation in one benchmark but low in another, indicating that our Q-matrix and ability taxonomy produce coherent, cross-dataset interpretable profiles.

All reported correlations with moderate or stronger strength have p-values < 0.01, confirming statistical significance. Together, these results validate that our method provides reliable, fine-grained diagnostics, provided sufficient items per dimension are available.

\subsection{Cross-Benchmark Consistency of Fine-Grained Ability Estimates}
\label{sec:validity}

Building upon the within-benchmark validity analysis, we further evaluate the generalization of our diagnostic framework across benchmarks with differing distributions of ability dimensions. To ensure reliable comparison, we retain only those ability dimensions for which both benchmarks contain more than 10 associated items. This threshold is chosen to mitigate instability from sparse item coverage.

\begin{figure}[htbp]
    \centering
    \resizebox{\linewidth}{!}{
        \includegraphics{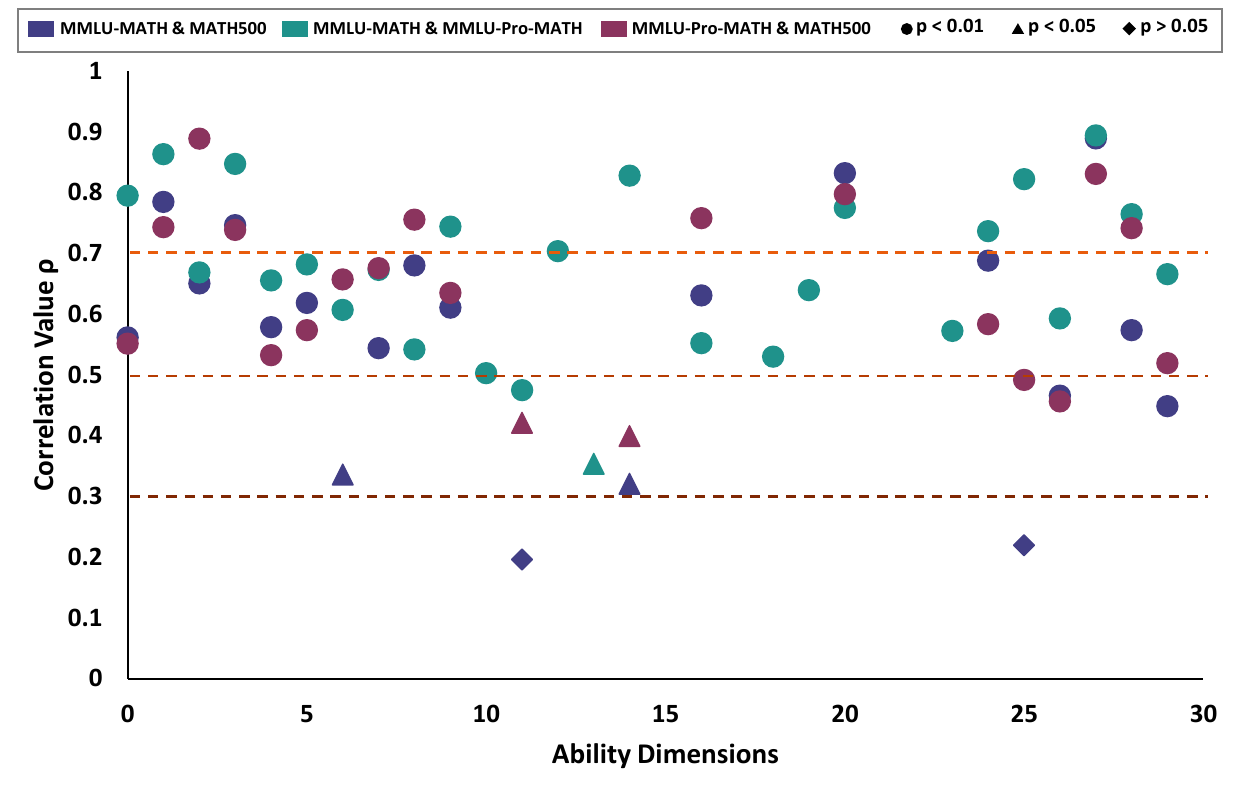}
    }
    \caption{
        Cross-benchmark correlation of estimated fine-grained ability levels. 
        Spearman's $\rho$ between paired benchmarks, computed over 19 of 35 ability dimensions with $>10$ items in all benchmarks. 
        Color: purple = MMLU-MATH vs MATH500, blue = MMLU-MATH vs MMLU-Pro-MATH, red = MMLU-Pro-MATH vs MATH500. 
        Shape: squares ($p < 0.01$), triangles ($p < 0.05$), diamonds ($p > 0.05$). 
        Dashed lines mark thresholds: $\rho = 0.3$ (weak), $0.5$ (moderate), $0.7$ (strong). 
        14 of these 19 dimensions show $\rho > 0.5$ across all pairwise comparisons.
    }
    \label{fig:diff_bench}
\end{figure}

We compute Spearman's rank correlation coefficient ($\rho$) between the estimated ability scores for each dimension across benchmark pairs. 
The results are visualized in Figure~\ref{fig:diff_bench}, with correlation strength categorized as follows: strong ($\rho > 0.7$), moderate ($0.5 < \rho \le 0.7$), and weak ($0.3 < \rho \le 0.5$). All reported correlations with $\rho > 0.3$ are statistically significant at $p < 0.05$; those with $\rho > 0.5$ satisfy $p < 0.01$.

Overall, the fine-grained ability estimates demonstrate strong cross-benchmark consistency.
Among the 19 ability dimensions that are valid across all three benchmarks (i.e., each has more than 10 associated items in every benchmark), 14 dimensions exhibit moderate or stronger correlation ($\rho > 0.5$) across all three benchmark pairs, indicating robust generalization. 
At the aggregate level, the median pairwise correlation across all dimension-benchmark combinations is $\rho = 0.66$, and 86\% of shared dimension-benchmark pairs achieve at least moderate correlation. 

These results confirm that our method produces stable, generalizable ability profiles across diverse mathematical benchmarks, reinforcing its suitability for evaluation.

\subsection{Predicting Model Performance on Unseen Items}
\label{sec:prediction}

Unlike indirect evaluation methods that rely solely on aggregate test set performance, our fine-grained ability evaluation framework, grounded in mIRT, enables prediction of correct or incorrect responses to unseen items based on the association between each item and a set of fine-grained abilities defined by the Q-matrix.

\begin{figure}
    \centering
    \resizebox{\linewidth}{!}{
        \includegraphics{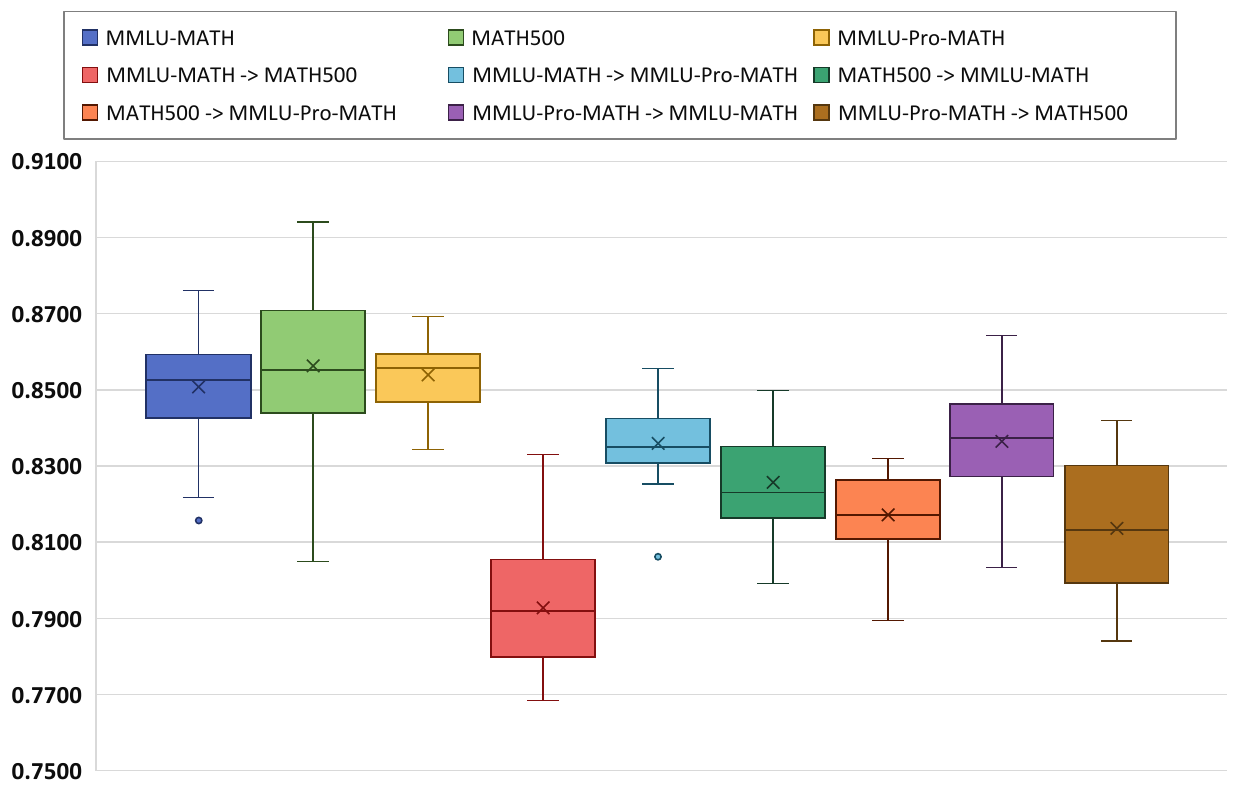}
    }
    \caption{
        AUC distribution across 41 LLMs for nine prediction scenarios. 
        Boxplots show median, IQR, and outliers (whiskers = 1.5$\times$IQR). 
        First three: within-benchmark prediction on MMLU-MATH, MATH500, MMLU-Pro-MATH; 
        remaining six: cross-benchmark prediction over all ordered benchmark pairs. 
        All scenarios significantly outperform baselines under identical item assumptions: 
        accuracy-based (AUC = 0.50), 
        standard IRT (AUC = 0.50), 
        and random prediction (mean AUC = 0.49).
    }
    \label{fig:prediction}
    
\end{figure}

We evaluate this predictive capability on three mathematical benchmarks: MMLU-MATH, MMLU-Pro-MATH, and MATH500. The area under the ROC curve (AUC) is used as the evaluation metric, as it is threshold-invariant and well-suited for binary response prediction in psychometric settings. For each scenario, we compute AUC separately for each of the 41 models, then report the distribution of these per-model AUC scores. Experiments are structured into two settings, with results visualized in Figure~\ref{fig:prediction}.

\textbf{(a) Within-Benchmark Prediction.}  
For each benchmark, we randomly split its items into an evaluation set (90\%) and a prediction set (10\%), repeating this procedure 10 times to account for partition variance. We estimate model ability levels using the evaluation set, then predict responses on the prediction set using the IRT-based scoring function (Section~\ref{sec:model}). For unseen items, difficulty and discrimination parameters are fixed to neutral defaults ($I_{\text{diff}} = 0.5$, $I_{\text{disc}} = 1.0$), following standard IRT conventions.

(b) \textbf{Cross-Benchmark Prediction.}  
We construct six ordered benchmark pairs (e.g., MMLU-MATH $\rightarrow$ MATH500), using the source benchmark for ability estimation and the target for prediction. Target items use the same fixed parameters ($I_{\text{diff}} = 0.5$, $I_{\text{disc}} = 1.0$).

We compare against three baselines under identical unseen-item conditions:  
(1) \textbf{Accuracy-based}: uses model's overall accuracy (AUC = 0.500);  
(2) \textbf{Standard unidimensional IRT}: models a single latent trait (AUC = 0.500);  
(3) \textbf{Random prediction}: samples probabilities from $\mathcal{N}(0.5, 0.2^2)$ truncated to $[0,1]$, yielding AUCs of 0.4968 $\pm$ 0.1157 (MMLU-MATH), 0.4708 $\pm$ 0.1326 (MMLU-Pro-MATH), and 0.4992 $\pm$ 0.0586 (MATH500).

Within-benchmark predictions achieve mean AUC > 0.81, while cross-benchmark results range from 0.77 to 0.81.
Results in Figure~\ref{fig:prediction} show that all scenarios significantly outperform these baselines, demonstrating that our framework reliably predicts model behavior on unseen items even across benchmarks with differing item--ability distributions.

\subsection{Cross-Domain Generalization of the Diagnostic Framework}

To evaluate the generalizability of our evaluation framework beyond mathematical reasoning, we extend the analysis to three additional domains: physics, chemistry, and computer science, using the corresponding subsets from the MMLU-Pro benchmark. Following the pipeline in Sections~\ref{sec:q}--\ref{sec:model}, we construct domain-specific Q-matrices via expert-supervised LLM annotation, aligning ability dimensions with each domain's knowledge structure.

For criterion validity, we analyze only those ability dimensions with sufficient item coverage (>10 associated items). The proportion of dimensions achieving moderate or stronger correlation ($\rho > 0.5$) is as follows:  
\textbf{Chemistry}: 48 out of 58 dimensions (82.8\%);  
\textbf{Physics}: 27 out of 27 dimensions (100\%);  
\textbf{Computer Science}: 10 out of 12 dimensions (83.3\%).

For prediction performance, the AUC values are 0.7664 in chemistry, 0.7925 in physics, and 0.7824 in computer science. These results fall within the range observed in mathematical benchmarks (AUC: 0.77--0.84), indicating that the framework maintains consistent predictive power when transferred to new scientific and technical domains. All cross-domain predictions significantly outperform baselines under identical conditions: accuracy-based (AUC = 0.500), standard unidimensional IRT (AUC = 0.500), and random prediction with mean AUC of 0.4927 $\pm$ 0.0838 (chemistry), 0.4958 $\pm$ 0.0894 (physics), and 0.5142 $\pm$ 0.1078 (computer science).

These findings confirm that our method can be effectively extended beyond mathematics through curriculum-aligned ability definition and Q-matrix construction, without loss of diagnostic or predictive fidelity.

%% file: sections/relate.tex
\section{Related Work}

\subsection{Large Language Model Evaluation}

Current evaluation of LLMs relies primarily on standardized benchmarks that report aggregate performance metrics. The MMLU benchmark~\cite{hendrycks2020measuring} evaluates cross-domain knowledge through multiple-choice questions spanning 57 subjects, while MATH and GSM8K~\cite{cobbe2021gsm8k} assess mathematical reasoning using specialized problem sets. More recent efforts such as Humanity's Last Exam~\cite{phan2025humanity} compile expert-level challenges to probe advanced deductive capabilities. Despite their widespread adoption, these benchmarks treat each task as a monolithic unit, collapsing diverse cognitive and knowledge-based processes into a single scalar score. This coarse-grained approach yields inconsistent model rankings across benchmarks and provides limited insight into the underlying sources of success or failure.

To address this limitation, several works have introduced more structured evaluation schemes. AtomThink~\cite{xiang2024atomthink} models complex reasoning as compositions of atomic operations, illustrating how higher-level capabilities emerge from basic modules. FineMath~\cite{liu2024finemath} decomposes elementary mathematics into 17 problem types to enable fine-grained analysis of reasoning pipelines. SciEval~\cite{sun2024scieval} proposes a four-dimensional taxonomy, including knowledge, application, analysis, and evaluation, to measure 24 scientific competencies.

However, these approaches rely on manually defined taxonomies that reflect designer-specific assumptions about ability structure. Moreover, they either assess abilities in isolation or focus on generative modeling rather than diagnostic evaluation. Crucially, none establish a shared, psychometrically grounded ability space that supports (1) cross-benchmark comparison under heterogeneous distributions, (2) interpretable diagnosis of model ability profiles, or (3) prediction of model behavior on unseen items.

\subsection{Psychometrics and Cognitive Diagnosis}

Psychometrics provides a formal framework for measuring latent traits such as cognitive abilities and domain knowledge through observable responses. Classical Test Theory (CTT)~\cite{devellis2006classical} models an observed score as the sum of a true ability and random error, but treats ability as a single global construct. Item Response Theory (IRT)~\cite{harvey1999item} models the probability of a correct response as a function of both examinee ability and item properties, such as difficulty and discrimination, enabling more precise individual assessment. However, standard IRT assumes a unidimensional latent trait, which is insufficient when tasks require multiple distinct competencies. mIRT~\cite{ackerman2014multidimensional} extends this paradigm by representing ability as a vector of fine-grained traits, allowing items to depend on multiple underlying dimensions simultaneously.

This advancement is formalized in Cognitive Diagnosis Models (CDMs)~\cite{de2009cognitive}, which explicitly define the relationship between test items and required skills via a Q-matrix. The Q-matrix is a binary mapping that indicates which attributes each item measures. CDMs such as DINA~\cite{junker2001cognitive} and G-DINA~\cite{de2011generalized} estimate mastery profiles over discrete skill sets, supporting both diagnostic interpretation and prediction of future responses.
Recent work integrates neural networks into this framework. NeuralCD~\cite{wang2023neuralcd} replaces linear link functions with deep architectures to model complex interactions between skills and items, while preserving the interpretability of the Q-matrix. These models demonstrate that structured psychometric frameworks can be enhanced with modern machine learning without sacrificing diagnostic transparency.

Collectively, these approaches establish a principled foundation for decomposing performance into interpretable ability components, a perspective increasingly relevant to the evaluation of LLMs.

\subsection{Diagnosis Model-based LLM Evaluation}

Recent work has begun to adapt psychometric models for LLM evaluation. Castleman et al.~\cite{castleman2025rethinking} analyze mathematical benchmarks and argue that datasets like GSM8K and MathOdyssey fail to reliably assess state-of-the-art models, advocating for IRT-based analysis to improve evaluation validity. Liu et al.~\cite{liu2023what} apply IRT to evaluate 51 recommendation algorithms, showing that high aggregate rankings do not necessarily reflect progress in user preference modeling. Zhou et al.~\cite{zhou2025lost} use IRT to construct neural networks that mitigate the impact of benchmark difficulty and complexity on model assessment. Rodriguez et al.~\cite{rodriguez-etal-2021-evaluation} employ IRT to quantify how item characteristics affect model rankings, though they do not estimate underlying abilities.

Building on these insights, JE-IRT~\cite{yao2025je} jointly embeds LLMs and items into a shared space using classical IRT, decomposing a single score into multiple ability dimensions. Similarly, Zhao et al.~\cite{zhao2025can} adopt a Bayesian framework to predict model performance in vision-language tasks. However, both methods require partial response records for target items to perform matrix completion, whereas our framework predicts items with no prior responses by mapping them into a shared ability space via the Q-matrix, enabling cross-benchmark generalization and interpretable diagnosis.

Despite these advances, current diagnosis-based evaluations remain limited in three key aspects. First, they lack a principled, fine-grained ability taxonomy; most define abilities post hoc or at the task level. Second, they are typically validated within a single benchmark, with little evidence of cross-benchmark or cross-domain generalization. Third, they focus on fitting observed responses rather than predicting performance on novel items. In contrast, our framework defines a unified 35-dimensional ability space spanning domain knowledge and cognitive processes, constructs a Q-matrix to map items across benchmarks, and demonstrates accurate prediction on unseen items both within and across scientific domains.

%% file: sections/conclusion.tex
\section{Conclusion and Future Directions}
\label{sec:conclusion}

We propose a cognitive diagnostic framework for fine-grained LLM evaluation grounded in multidimensional Item Response Theory (mIRT). The framework constructs domain-specific ability taxonomies by integrating cognitive theory with domain knowledge structures; for mathematics, this yields a 35-dimensional taxonomy spanning domain knowledge and cognitive processes. Evaluated on 41 models, the framework demonstrates strong criterion validity, consistent ability estimates across benchmarks, and accurate prediction of unseen items with AUC ranging from 0.80 to 0.89 within benchmarks and from 0.77 to 0.86 across benchmarks, substantially exceeding trivial baselines.
The framework generalizes across scientific domains, producing consistent diagnostic performance in physics (27 dimensions), chemistry (58 dimensions), and computer science (12 dimensions). By estimating interpretable fine-grained ability profiles, our approach provides a principled foundation for LLM assessment with potential to inform targeted training, ability-guided model selection, ability-aware benchmark design, and alignment between benchmark evaluations and real-world task demands.

Limitations include the manual effort required for taxonomy construction and Q-matrix annotation, potential subjectivity in mapping items to fine-grained abilities, and uneven ability coverage in existing benchmarks. Future work will extend the framework to additional domains, develop ability-aware benchmarks with balanced coverage, and explore ability-guided model training based on diagnostic profiles.

%% file: appendix/prompt.tex
\clearpage
\section{Methodology for Fine-Grained Ability Taxonomy and Q-Matrix Construction}
\label{app:prompt}

We employ a human-in-the-loop framework that leverages large language models (LLMs) under expert supervision to construct a fine-grained ability taxonomy grounded in domain knowledge and Bloom's taxonomy, and to subsequently derive the Q-matrix. The process consists of three sequential stages. First, in item analysis, we map benchmark items to relevant cognitive attributes using established educational frameworks (e.g., K-16 mathematics curricula) and Bloom's taxonomy, yielding an initial attribute pool; human experts sample LLM-generated mappings to assess validity and iteratively refine the prompting strategy to improve alignment with pedagogical standards. Second, during taxonomy construction, we organize this pool into a hierarchical structure: the LLM proposes groupings and provides concise definitions for abilities at each level, which domain experts then fully review to ensure coherence, non-redundancy, and fidelity to educational theory. Third, in Q-matrix construction, we select an appropriate granularity level from the taxonomy as the target ability dimensions for evaluation, and under expert supervision, the LLM annotates each benchmark item with binary indicators denoting required abilities, producing the final Q-matrix.
The prompts used in each stage are provided below. 

\subsection{Prompt of Item Analysis}
In the item analysis stage, we prompt the LLM to map each benchmark item to relevant cognitive attributes based on domain knowledge and Bloom's taxonomy. The prompt used is as follows.

\textbf{System Prompt:}
\begin{lstlisting}
You are a professional mathematics education assessment expert with 15 years of experience in mathematics curriculum design and test question analysis. You excel at precisely deconstructing the ability structure required for test questions from the dual perspectives of cognitive psychology and mathematics pedagogy.
Please analyze the following mathematics test question and output **both the specific abilities covered by this question and the reasons why these abilities are required**.
For each type of ability, please include:
1. **What specific abilities are covered** - List the exact abilities needed
2. **Why these abilities are covered** - Explain how each ability is applied in solving this specific question

**Ability Categories:**
- Knowledge-based abilities: Identify specific, concrete mathematical facts, concepts, formulas, theorems, or procedures that the user must recall or recognize. Output must be extremely concise, but not too closely related to the specific formulas or data in the question. Using only standard domain-specific keywords or very short phrases (e.g., 'Pythagorean theorem', 'quadratic formula', 'derivative rules'). **Avoid complete sentences or explanatory clauses.**
- Cognitive abilities: Identify the higher-order mental processes and strategic abilities required to apply the knowledge. Output must be concise, using generic, transferable skill names expressed as short noun phrases or gerunds (e.g., 'algebraic manipulation', 'logical reasoning', 'modeling', 'interpretation'). **Avoid broad verbs like 'solve' or 'use'; strive for precise, compact terms.**

**Strict Output Requirements:**
- Each ability must be directly relevant to answering the test question
- Knowledge-based abilities should be specific to mathematical concepts
- Cognitive abilities should use professional ability names (e.g., "logical reasoning," "spatial imagination")
- For each ability, provide a clear explanation of why it's needed for this specific question
- Only analyze abilities and their justifications; do not evaluate difficulty or provide suggestions
- Output strictly in JSON format, without any additional text
- All skills must be in English, using precise and commonly accepted terminology

**Output Format:**
{
  "knowledge_abilities": [
    {
      "ability": "mathematical knowledge point 1",
      "explanation": "why this knowledge is required for solving this question"
    },
    {
      "ability": "specific mathematical knowledge point 2",
      "explanation": "why this knowledge is required for solving this question"
    }
  ],
  "cognitive_abilities": [
    {
      "ability": "cognitive ability name 1",
      "explanation": "why this cognitive ability is applied in solving this question"
    },
    {
      "ability": "cognitive ability name 2",
      "explanation": "why this cognitive ability is applied in solving this question"
    }
  ]
}
\end{lstlisting}

\textbf{User Prompt:}
\begin{lstlisting}
Analyze the following problem:
[problem text]
\end{lstlisting}

\subsection{Prompt of Taxonomy Construction}
In the taxonomy construction stage, human experts provide an initial set of fine-grained knowledge points derived from item analysis. The LLM is then tasked with organizing these into a hierarchical competency framework grounded in educational theory. The prompts used are as follows.

\textbf{System Prompt:}
\begin{lstlisting}
You are an expert in mathematics education and curriculum design. Your task is to construct a comprehensive and well-structured mathematical competency framework based on a set of fine-grained knowledge points provided by the user.

**1. Core Design Principles**
- **Theoretical Anchoring**: The framework design must be primarily based on the  **Revised Bloom's Taxonomy**'s cognitive process dimension (Remember, Understand, Apply, Analyze, Evaluate, Create) and knowledge dimension (Factual, Conceptual, Procedural, Metacognitive). Additionally, references can be made to classifications in mathematics education, such as  **Declarative Knowledge**  (related to abstraction and logical reasoning) and  **Procedural Knowledge**  (related to computation and modeling).
- **Granularity Consistency**: This is a key requirement. Each branch of the framework must maintain balanced hierarchical depth and subdivision granularity.
    - **Horizontal Consistency**: All nodes at the same level (e.g., all secondary nodes) should represent competency domains of roughly the same scale.
    - **Vertical Progression**: The division of levels should have clear logic (e.g., "Knowledge Domain -> Core Concept -> Specific Component" or "Cognitive Process -> Specific Practice -> Performance Description"), with a depth typically not exceeding 4-5 layers to avoid some paths being too deep while others are too shallow.

**2. Competency Framework Structure Requirements**
The framework must consist of the following two parts:
- **Knowledge Abilities**: Reflects familiarity with and mastery of mathematical knowledge itself. The structure should embody the internal logic of mathematical content (e.g., Number and Quantity, Geometry, etc.).
- **Cognitive Abilities**: Reflects the application and thinking processes applied to mastered knowledge. The structure should reflect a spectrum of cognitive skills from lower-order to higher-order, integrating mathematical practices such as problem-solving, reasoning, modeling, and communication.

**3. Level Definition and Description Specifications**
You must provide a clear description for  **every custom non-leaf node**  (i.e., all category and subcategory nodes) in the framework to ensure the intent of the hierarchical construction is explicit and aligns with research consensus. Descriptions should follow this format:
- `"level_definition"`: Define the core connotation of the abilities at this level in 1-2 sentences.
- `"differentiation_criteria"`: Explain the principles (e.g., different knowledge domains, cognitive behaviors, practical contexts) used to differentiate between sibling nodes under the same parent node.
- `"example_verbs_or_actions"`: Provide 2-3 typical action verbs or task examples that align with the abilities at this level (e.g., for an "Apply" level, these could be "execute a computation", "use a formula to solve a routine problem").

**4. Knowledge Point Mapping and Output Format**
- You must ensure that all  **fine-grained knowledge points**  provided by the user can be accurately categorized as  **leaf nodes**  under the appropriate parent node at the most granular level of the framework.
- The output must be in strict, directly parsable JSON format, containing no non-ASCII characters. The JSON structure must strictly adhere to the following template, with your "description" for each non-leaf node embedded in the corresponding node's  `"description"`  field.

**Output Format:**
{
  "knowledge_abilities": {
    "children": {
      "node1": {
        "name": "node1",
        "description": "description of node1",
        "children": {
          "node1-1": {
            "name": "node11",
            "description": "description of node1-1",
            "children": {
              "node1-1-1": {
                "name": "node111",
                "description": "description of node1-1-1",
                "children": {
                  "node1-1-1-1": "knowledge point 11",
                  "node1-1-1-2": "knowledge point 12"
                }
              }
            }
          }
        }
      }
    }
  },
  "cognitive_abilities": {
    "children": {
      "node2": {
        "name": "node2",
        "description": "description of node2",
        "children": {
          "node2-1": {
            "name": "node21",
            "description": "description of node2-1",
            "children": {
              "node2-1-1": {
                "name": "node211",
                "description": "description of node2-1-1",
                "children": {
                  "node2-1-1-1": "knowledge point 21",
                  "node2-1-1-2": "knowledge point 22"
                }
              }
            }
          }
        }
      }
    }
  }
}

Note: Ensure all provided knowledge points can be found with corresponding parent nodes in the tree structure, and place them at appropriate levels. The output must be valid JSON that can be directly parsed.
\end{lstlisting}

\textbf{User Prompt:}
\begin{lstlisting}
Please generate the competency framework based on the following knowledge points:
[selected fine-grained abilities]
\end{lstlisting}

\subsection{Prompt of Q-Matrix Construction}
In the Q-matrix construction stage, the LLM annotates each benchmark item by mapping it to relevant abilities from the finalized taxonomy. The system and user prompts used are as follows.

\textbf{System Prompt:}
\begin{lstlisting}
You are a professional math test item analysis assistant. Please strictly adhere to the following requirements for the analysis task:

**Task Requirements:**
1. For each input test question, analyze the mathematical abilities it assesses.
2. **Abilities** must and can only come from the user-provided ability list. Do not add points outside the list.
3. The output must be in **strict JSON format**, with the specific structure detailed below.
4. For each related ability, provide a brief explanatory note justifying why and how the question assesses this ability.

**Analysis Principles:**
1. Read each question carefully to identify the core concepts, formulas, theorems, or methods being tested.
2. Strictly match the identified core concepts against the provided abilities list.
3. When matching, the ability name should be an exact or core synonymous match (as per the given list).
4. Explanations should be specific, pointing to conditions, problems, or solution steps within the question.
5. **All special characters must be expressed in escaped form** to ensure valid JSON format. For example:
   - Double quotes should be escaped as \"
   - Backslashes should be escaped as \\
   - Other JSON special characters should be escaped accordingly
6. **JSON key names must strictly adhere to the following specifications**:
   - Key names **must not contain any extra spaces**
   - Key names must exactly match those in the example.

**JSON Key Examples:** "question_content", "related_abilities", "ability", "explanation"

**Output JSON Format:**
{
  "question_content": "Given a right triangle with legs of lengths 3 and 4, find the length of the hypotenuse.",
  "related_abilities": [
    {
      "ability": "Pythagorean Theorem",
      "explanation": "The question explicitly gives the two legs of a right triangle and asks for the hypotenuse, requiring direct application of the Pythagorean Theorem \"a^2 + b^2 = c^2\"."
    },
    {
      "ability": "Square Root Computation",
      "explanation": "After applying the Pythagorean Theorem, the solver must compute the square root of 25 to obtain the final answer."
    }
  ]
}

**Output Format Specification:**
- Use the above JSON structure. Place the analysis results for each question in a list if multiple questions are processed (though typically one at a time in our pipeline).
\end{lstlisting}

\textbf{User Prompt:}
\begin{lstlisting}
Analyze the following problem:
[problem text]
\end{lstlisting}

%% file: appendix/taxonomy.tex
\clearpage

\section{Mathematical Ability Taxonomy}
\label{app:taxonomy}

Following our methodology for constructing fine-grained ability taxonomies grounded in cognitive theory, we analyze three mathematical benchmarks (MMLU-MATH, MMLU-Pro-MATH, and MATH500) through the lens of K-16 mathematics curricula and Bloom's taxonomy. This yields a hierarchical taxonomy for mathematical problem solving.

The final taxonomy comprises 35 fine-grained abilities, refined from an initial set of 45 dimensions through iterative expert review. These abilities are organized into two primary dimensions: \textit{Knowledge} (abilities 0--21) and \textit{Cognitive} (abilities 22--34). The structure is visualized in Figure~\ref{fig:finegrain}, with the complete numbered list presented below.

\subsection{Knowledge Dimension}
\begin{itemize}
    \item \textbf{Arithmetic}
        \begin{itemize}
            \item \textit{Properties of Numbers (ability\_0)}: Recognize characteristics such as parity, divisibility, and magnitude.
            \item \textit{Operations of Numbers (ability\_1)}: Perform computations on integers, fractions, and decimals.
            \item \textit{Number Theory (ability\_2)}: Apply concepts of primes, divisibility, and modular arithmetic.
        \end{itemize}
    
    \item \textbf{Algebra}
        \begin{itemize}
            \item \textit{Expressions and Equations (ability\_3)}: Simplify symbolic expressions and solve algebraic equations.
            \item \textit{Polynomials (ability\_4)}: Manipulate polynomials through addition, multiplication, and factorization.
        \end{itemize}
    
    \item \textbf{Functions}
        \begin{itemize}
            \item \textit{Concepts of Functions (ability\_5)}: Understand input-output relationships and function properties.
            \item \textit{Elementary Functions (ability\_6)}: Work with linear, quadratic, exponential, and logarithmic functions.
        \end{itemize}
    
    \item \textbf{Geometry}
        \begin{itemize}
            \item \textit{Euclidean Geometry (ability\_7)}: Apply axioms and theorems to plane geometric figures.
            \item \textit{Analytic Geometry (ability\_8)}: Use coordinate systems to represent and solve geometric problems.
            \item \textit{Geometric Measurement (ability\_9)}: Compute lengths, areas, volumes, and angles.
        \end{itemize}
    
    \item \textbf{Calculus}
        \begin{itemize}
            \item \textit{Limits and Continuity (ability\_10)}: Analyze function behavior near points or at infinity.
            \item \textit{Differential Calculus (ability\_11)}: Compute derivatives and apply them to rates of change and optimization.
            \item \textit{Integral Calculus (ability\_12)}: Evaluate integrals and interpret them as accumulated quantities.
        \end{itemize}
    
    \item \textbf{Probability and Statistics}
        \begin{itemize}
            \item \textit{Descriptive Statistics (ability\_13)}: Summarize data using measures of center and spread.
            \item \textit{Probability (ability\_14)}: Calculate likelihoods using classical or theoretical models.
            \item \textit{Statistical Inference (ability\_15)}: Draw conclusions about populations from sample data.
        \end{itemize}
    
    \item \textbf{Linear Algebra}
        \begin{itemize}
            \item \textit{Vectors and Matrices (ability\_16)}: Perform operations including addition and multiplication.
            \item \textit{Linear Systems (ability\_17)}: Solve systems of linear equations using algebraic or matrix methods.
            \item \textit{Vector Spaces (ability\_18)}: Analyze subspaces, bases, and dimension.
        \end{itemize}
    
    \item \textbf{Discrete Mathematics}
        \begin{itemize}
            \item \textit{Sets and Logic (ability\_19)}: Apply set operations and logical reasoning.
            \item \textit{Combinatorial Counting (ability\_20)}: Count arrangements using permutations and combinations.
            \item \textit{Graph Theory (ability\_21)}: Analyze connectivity and structures in networks.
        \end{itemize}
\end{itemize}

\subsection{Cognitive Dimension}
\begin{itemize}
    \item \textbf{Information Recognition and Extraction}
        \begin{itemize}
            \item \textit{Element Identification (ability\_22)}: Recognize and classify mathematical objects such as numbers or symbols.
            \item \textit{Knowledge Retrieval (ability\_23)}: Recall definitions, formulas, or procedures from memory.
        \end{itemize}
    
    \item \textbf{Concept Understanding and Manipulation}
        \begin{itemize}
            \item \textit{Interpretation and Connection (ability\_24)}: Link mathematical representations to meanings or prior knowledge.
            \item \textit{Concept Manipulation (ability\_25)}: Reorganize or transform concepts for new problem contexts.
        \end{itemize}
    
    \item \textbf{Analysis, Synthesis, and Verification}
        \begin{itemize}
            \item \textit{Analysis and Decomposition (ability\_26)}: Break complex problems into simpler components.
            \item \textit{Synthesis and Evaluation (ability\_27)}: Combine ideas and judge solution validity or efficiency.
        \end{itemize}
    
    \item \textbf{Critical Reasoning and Debugging}
        \begin{itemize}
            \item \textit{Error Diagnosis (ability\_28)}: Detect and explain mistakes in reasoning or computation.
            \item \textit{Robustness Analysis (ability\_29)}: Test solutions under varying conditions or edge cases.
        \end{itemize}
    
    \item \textbf{Creative Generation}
        \begin{itemize}
            \item \textit{Generation and Construction (ability\_30)}: Create new mathematical objects, models, or representations.
            \item \textit{Abstraction and Generalization (ability\_31)}: Derive general principles from specific examples.
        \end{itemize}
    
    \item \textbf{Multi-step Problem Solving}
        \begin{itemize}
            \item \textit{Procedure Construction (ability\_32)}: Design step-by-step methods for novel problems.
            \item \textit{Procedure Execution (ability\_33)}: Carry out algorithms or computational steps accurately.
            \item \textit{Conventional Modeling (ability\_34)}: Translate real-world scenarios into mathematical forms.
        \end{itemize}
\end{itemize}

%% file: appendix/domains.tex
\clearpage

\section{Extended Domain Taxonomies}
\label{app:extended_taxonomies}

To demonstrate the generalizability of our taxonomy construction methodology, we apply it to three additional scientific domains beyond mathematics: physics, chemistry, and computer science. Each domain-specific taxonomy is constructed through expert-supervised LLM annotation aligned with the field's knowledge structure. Below we present the resulting ability taxonomies.

\subsection{Physics}
\label{app:physics}

The physics taxonomy comprises 27 fine-grained abilities spanning classical mechanics, electromagnetism, thermodynamics, optics, modern physics, astronomy, and cognitive processes. The complete list is presented below.

\begin{itemize}
    \item \textit{Kinematics and Dynamics (ability\_0)}: Analyze motion and forces using Newtonian principles.
    \item \textit{Rotational and Oscillatory Motion (ability\_1)}: Apply concepts of torque, angular momentum, and harmonic oscillation.
    \item \textit{Gravitation and Orbital Mechanics (ability\_2)}: Model gravitational interactions and orbital trajectories.
    \item \textit{Wave Basics (ability\_3)}: Describe wave properties including frequency, wavelength, and propagation.
    \item \textit{Geometric Optics (ability\_4)}: Apply ray tracing and lens/mirror equations to optical systems.
    \item \textit{Physical Optics (ability\_5)}: Analyze interference, diffraction, and polarization phenomena.
    \item \textit{Temperature and Units (ability\_6)}: Convert between temperature scales and thermal units.
    \item \textit{Heat, Calorimetry, and Phase (ability\_7)}: Compute heat transfer and analyze phase transitions.
    \item \textit{Gas Laws, Processes, and Radiation (ability\_8)}: Apply ideal gas law and analyze thermodynamic cycles.
    \item \textit{Electrostatics and Fields (ability\_9)}: Calculate electric fields and potentials from charge distributions.
    \item \textit{Magnetism and Induction (ability\_10)}: Analyze magnetic fields and electromagnetic induction.
    \item \textit{Circuits and Electromagnetic Waves (ability\_11)}: Solve circuit problems and describe electromagnetic wave propagation.
    \item \textit{Quantum and Atomic Physics (ability\_12)}: Explain quantization, atomic structure, and wave-particle duality.
    \item \textit{Nuclear Physics and Radioactivity (ability\_13)}: Analyze nuclear reactions, decay processes, and radiation.
    \item \textit{Relativity and High-Energy Physics (ability\_14)}: Apply special relativity principles and describe particle interactions.
    \item \textit{Solar System Dynamics (ability\_15)}: Model planetary motion and gravitational interactions within the solar system.
    \item \textit{Observational Astronomy (ability\_16)}: Interpret astronomical observations and measurement techniques.
    \item \textit{Planetary Geology and Atmospheres (ability\_17)}: Analyze surface features and atmospheric properties of celestial bodies.
    \item \textit{Units and Conversions (ability\_18)}: Perform dimensional conversions across physical quantities.
    \item \textit{Dimensional Analysis and Estimation (ability\_19)}: Derive relationships and estimate magnitudes using dimensional reasoning.
    \item \textit{Vectors and Calculus Tools (ability\_20)}: Apply vector operations and calculus to physical problems.
    \item \textit{Interferometry and Zone Relations (ability\_21)}: Interpret interference patterns and Fresnel zone constructions.
    \item \textit{Lens Design and Cardinal Points (ability\_22)}: Characterize optical systems using principal planes and focal points.
    \item \textit{Factual Knowledge (ability\_23)}: Recall fundamental physics facts, constants, and definitions.
    \item \textit{Conceptual Understanding (ability\_24)}: Explain physical principles and relationships between concepts.
    \item \textit{Procedural Skill (ability\_25)}: Execute problem-solving procedures and mathematical derivations.
    \item \textit{Metacognitive Awareness (ability\_26)}: Monitor and regulate problem-solving strategies and solution validity.
\end{itemize}

\subsection{Chemistry}
\label{app:chemistry}

The chemistry taxonomy comprises 58 fine-grained abilities spanning core concepts, physical chemistry, inorganic chemistry, organic chemistry, analytical techniques, and cognitive processes. The complete list is presented below.

\begin{itemize}
    \item \textit{Units and Conversions (ability\_0)}: Perform dimensional conversions across chemical contexts.
    \item \textit{Physical Constants and Reference Values (ability\_1)}: Recall and apply fundamental constants in chemical calculations.
    \item \textit{Atomic Structure and Electron Configuration (ability\_2)}: Describe atomic models and electron arrangements.
    \item \textit{Chemical Bonding and Intermolecular Forces (ability\_3)}: Analyze covalent, ionic, and intermolecular interactions.
    \item \textit{Ideal and Real Gases (ability\_4)}: Apply gas laws to ideal and non-ideal gas behavior.
    \item \textit{Thermodynamic Quantities and Processes (ability\_5)}: Calculate energy changes in chemical processes.
    \item \textit{Equilibrium Concepts (ability\_6)}: Analyze dynamic equilibrium in chemical systems.
    \item \textit{Acid-Base and p-Scales (ability\_7)}: Interpret acid-base behavior and pH calculations.
    \item \textit{Buffers and Titrations (ability\_8)}: Design and analyze buffer systems and titration curves.
    \item \textit{Potentials and Nernst (ability\_9)}: Calculate electrochemical potentials using Nernst equation.
    \item \textit{Faraday Laws and Charge Accounting (ability\_10)}: Relate electrical charge to chemical change in electrochemistry.
    \item \textit{Cell Notation and Conventions (ability\_11)}: Interpret standard electrochemical cell diagrams.
    \item \textit{Rate Laws and Integrated Forms (ability\_12)}: Determine reaction rates and integrated rate equations.
    \item \textit{Mechanisms and Approximations (ability\_13)}: Analyze reaction mechanisms and kinetic approximations.
    \item \textit{Hydrogenic Models and Energy Levels (ability\_14)}: Apply quantum models to atomic energy systems.
    \item \textit{Selection Rules and Photon Relations (ability\_15)}: Interpret spectroscopic transitions and photon interactions.
    \item \textit{Uncertainty and Time-Energy Relations (ability\_16)}: Apply quantum uncertainty principles in chemical contexts.
    \item \textit{Magnetic Resonance (EPR/NMR) (ability\_17)}: Analyze molecular structure using spectroscopic techniques.
    \item \textit{Optical Absorption (ability\_18)}: Interpret absorption spectra and Beer-Lambert law applications.
    \item \textit{Crystal Structures and Geometry (ability\_19)}: Describe crystalline arrangements and symmetry.
    \item \textit{Lattice Parameters and Distances (ability\_20)}: Calculate unit cell dimensions and interatomic distances.
    \item \textit{Crystal Density and Formula Units (ability\_21)}: Determine crystal densities from structural data.
    \item \textit{Reaction Stoichiometry and Balancing (ability\_22)}: Balance chemical equations and apply stoichiometric ratios.
    \item \textit{Mass-Mole-Amount Conversions (ability\_23)}: Convert between mass, moles, and particle quantities.
    \item \textit{Precipitation and Solubility Rules (ability\_24)}: Predict precipitation and solubility behavior.
    \item \textit{Concentration Measures (ability\_25)}: Calculate and convert between concentration units.
    \item \textit{Colligative Properties (ability\_26)}: Analyze vapor pressure, boiling point, and freezing point changes.
    \item \textit{Decay Laws and Activity (ability\_27)}: Apply radioactive decay kinetics to nuclear processes.
    \item \textit{Particle Properties (ability\_28)}: Characterize subatomic particles and their interactions.
    \item \textit{Functional Groups and Nomenclature (ability\_29)}: Identify organic functional groups and apply IUPAC naming.
    \item \textit{Representative Reactions (ability\_30)}: Recognize characteristic organic reaction types.
    \item \textit{Biomolecules and Properties (ability\_31)}: Analyze structures and functions of biological macromolecules.
    \item \textit{Toxicology and Safety (ability\_32)}: Evaluate chemical hazards and safety protocols.
    \item \textit{Notation and Precision (ability\_33)}: Apply significant figures and precision in chemical reporting.
    \item \textit{Stoichiometric and Equilibrium Schemas (ability\_34)}: Construct reaction schematics for complex systems.
    \item \textit{Factual Knowledge (ability\_35)}: Recall fundamental chemical facts and constants.
    \item \textit{Conceptual Understanding (ability\_36)}: Explain chemical principles and relationships.
    \item \textit{Procedural Skill (ability\_37)}: Execute chemical problem-solving procedures.
    \item \textit{Conceptual Interpretation (ability\_38)}: Interpret conceptual relationships in chemical systems.
    \item \textit{Representation Translation (ability\_39)}: Convert between symbolic, graphical, and verbal representations.
    \item \textit{Summarization and Categorization (ability\_40)}: Synthesize information and classify chemical concepts.
    \item \textit{Quantitative Computation and Unit Management (ability\_41)}: Perform numerical calculations with proper unit handling.
    \item \textit{Stoichiometric Reasoning and Bookkeeping (ability\_42)}: Track material balances in chemical processes.
    \item \textit{Modeling and Equation Setup (ability\_43)}: Formulate mathematical models for chemical phenomena.
    \item \textit{Problem Decomposition and Constraint Analysis (ability\_44)}: Break down complex problems and identify constraints.
    \item \textit{Comparative and Sign-Sensitive Reasoning (ability\_45)}: Compare chemical behaviors and interpret sign-dependent phenomena.
    \item \textit{Pattern Recognition and Trend Analysis (ability\_46)}: Identify patterns and trends in chemical data.
    \item \textit{Assumption and Approximation Management (ability\_47)}: Evaluate and justify modeling assumptions.
    \item \textit{Option Screening and Decision (ability\_48)}: Select appropriate methods and evaluate alternatives.
    \item \textit{Error Checking and Propagation (ability\_49)}: Verify results and analyze error propagation.
    \item \textit{Model Building and Equation Synthesis (ability\_50)}: Construct and synthesize chemical models.
    \item \textit{Multi-step Planning and Integration (ability\_51)}: Design integrated solution pathways for complex problems.
    \item \textit{Algorithmic and Computational Fluency (ability\_52)}: Execute computational procedures efficiently.
    \item \textit{Problem Solving and Strategy (ability\_53)}: Apply strategic approaches to chemical problem-solving.
    \item \textit{Reasoning and Algebra (ability\_54)}: Apply algebraic reasoning to chemical problems.
    \item \textit{Data Handling and Extraction (ability\_55)}: Process and extract information from chemical data.
    \item \textit{Unit Consistency and Dimensional Reasoning (ability\_56)}: Ensure dimensional consistency in calculations.
    \item \textit{Decision and Option Management (ability\_57)}: Evaluate and manage decision options in chemical contexts.
\end{itemize}

\subsection{Computer Science}
\label{app:cs}

The computer science taxonomy comprises 12 fine-grained abilities spanning discrete foundations, data systems, algorithmic reasoning, and professional practice. The complete list is presented below.

\begin{itemize}
    \item \textit{Discrete Mathematics (ability\_0)}: Apply logic, set theory, combinatorics, and graph theory to computational problems.
    \item \textit{Data Representation and Formats (ability\_1)}: Encode and interpret data using appropriate structures and encoding schemes.
    \item \textit{Data Management and Query (ability\_2)}: Design and query databases using relational or non-relational paradigms.
    \item \textit{Data Analysis and Machine Learning (ability\_3)}: Analyze datasets and apply supervised or unsupervised learning techniques.
    \item \textit{Societal Impact and Policy (ability\_4)}: Evaluate broader societal implications, legal frameworks, and policy considerations of computing technologies.
    \item \textit{Algorithmic Problem Solving (ability\_5)}: Design and analyze algorithms for computational tasks with attention to correctness and efficiency.
    \item \textit{Abstraction and Modeling (ability\_6)}: Construct abstract representations of real-world problems for computational solution.
    \item \textit{Data Reasoning (ability\_7)}: Infer patterns, relationships, and insights from structured or unstructured data.
    \item \textit{Systems Thinking (ability\_8)}: Analyze interactions between components in complex computational systems.
    \item \textit{Debugging and Testing (ability\_9)}: Identify, isolate, and resolve errors through systematic testing and validation.
    \item \textit{Communication and Collaboration (ability\_10)}: Articulate technical concepts and collaborate effectively in team settings.
    \item \textit{Ethical Decision Making (ability\_11)}: Apply ethical frameworks to navigate dilemmas in software development and deployment.
\end{itemize}

%% file: appendix/setup.tex
\clearpage

\section{Experiment Setup Details}
\label{app:setup}

We evaluate 41 language models on three mathematical benchmarks (MMLU-MATH, MMLU-Pro-MATH, and MATH500). For each model, we derive a 35-dimensional ability profile using multidimensional item response theory (mIRT) with the Q-matrix defined in Appendix B. The validity of these profiles is assessed along three dimensions: (1) construct validity through internal consistency, (2) cross-benchmark reliability, and (3) predictive utility for held-out item performance. Additionally, we evaluate cross-domain generalization by applying the same profiling pipeline to three non-mathematical subsets of MMLU-Pro (Chemistry, Physics, and Computer Science).

\subsection{Evaluated Models}
We evaluate a total of 41 language models across two categories: online API models and open-source models. The full list is presented in Table \ref{tab:evaluated-models}.

\input{table/evaluated-model.tex}

\subsection{Evaluated Benchmarks}
We evaluate models on six subsets across three benchmark suites: MMLU, MMLU-Pro, and MATH500. The full list is presented in Table \ref{tab:evaluated-benchmarks}.

\input{table/evaluated-benchmarks.tex}
We quantify the coverage of each of the 35 fine-grained abilities across the three mathematical benchmarks (MMLU-MATH, MMLU-Pro-MATH, MATH500) by counting the number of items associated with each ability in the Q-matrix. Table~\ref{tab:ability-coverage} reports both absolute counts and relative proportions (radio = count / total items per benchmark).

\subsection{Ability Coverage Statistics of Mathematical Benchmarks}
\label{app:coverage}

As shown in Table \ref{tab:ability-coverage}, we analysis the item coverage of each ability across three mathematical benchmarks.
\input{table/distirbution.tex}

\subsection{Correlation Analysis}
\label{app:corr}

We adopt the Spearman rank correlation coefficient ($\rho$) as the primary measure of association for three reasons. First, ability estimates from multidimensional item response theory (mIRT) are interval-scale but may violate the linearity and normality assumptions required by Pearson correlation. Second, our focus is on \textit{rank-order consistency} of model capabilities (e.g., whether a model consistently ranks higher across abilities), which Spearman directly captures. Third, Spearman correlation is robust to outliers and monotonic nonlinearities, critical given the heterogeneous performance distributions across 41 models and 35 ability dimensions.

To assess the validity and consistency of fine-grained ability profiles, we compute pairwise Spearman rank correlations between model ability vectors across the 35 mathematical dimensions.

\paragraph{Ability-Performance Association.}
For each ability dimension $i$, we compute the Spearman correlation between models' estimated abilities on dimension $i$ and their observed accuracy on items associated with that dimension. The observed accuracy of $m$-th model for dimension $i$ on a benchmark is calculated as:
\begin{equation}
    \label{eq:acc}
    \mathrm{ACC}_{mi} = \frac{\sum_{j=1}^N q_{ji} \cdot r_{jm}}{\sum_{j=1}^N q_{ji}},
\end{equation}
where:
\begin{itemize}
    \item $q_{ji} \in \{0, 1\}$ is the Q-matrix entry indicating whether item $j$ requires ability $i$,
    \item $r_{jm} \in \{0, 1\}$ is the correctness of $m$-th model's binary response on item $j$,
    \item $N$ is the total number of items in the benchmark.
\end{itemize}
This metric represents the proportion of correctly answered items among those that load on ability $i$.

\paragraph{Cross-Benchmark Consistency.}
For each model, we compute the Spearman correlation between its ability profiles estimated from different mathematical benchmarks. High correlation indicates stable rank-order ability estimation across benchmark pairs.

%% file: table/evaluated-model.tex
\begin{table}[htbp]
\centering
\small
\caption{List of evaluated language models with unique identifiers.}
\label{tab:evaluated-models}
\begin{tabular}{ccclc}
\toprule
\textbf{ID} & \textbf{Category} & \textbf{Series} & \textbf{Model} & \textbf{Parameter Size} \\
\midrule
1 & \multirow{6}{*}{Online Model} 
  & \multirow{2}{*}{DeepSeek} & deepseek-v3.2 & N/A \\
2 & & & deepseek-v3   & N/A \\
\cmidrule{3-5}
3 & & \multirow{4}{*}{GPT} & gpt-4o        & N/A \\
4 & & & gpt-5.1       & N/A \\
5 & & & gpt-o1        & N/A \\
6 & & & gpt-o3        & N/A \\
\midrule
7  & \multirow{36}{*}{Open Source Model} 
   & \multirow{3}{*}{Gemma3} & gemma3-12b-it         & 12B \\
8  & & & gemma3-1b-it          & 1B \\
9  & & & gemma3-4b-it          & 4B \\
\cmidrule{3-5}
10 & & GPT & gpt-oss-20b           & 20B \\
\cmidrule{3-5}
11 & & \multirow{2}{*}{Llama3} & Llama3-8B-Instruct    & 8B \\
12 & & & Llama3-8B             & 8B \\
\cmidrule{3-5}
13 & & \multirow{2}{*}{Llama3.1} & Llama3.1-8B-Instruct  & 8B \\
14 & & & Llama3.1-8B           & 8B \\
\cmidrule{3-5}
15 & & \multirow{4}{*}{Llama3.2} & Llama3.2-1B-Instruct  & 1B \\
16 & & & Llama3.2-1B           & 1B \\
17 & & & Llama3.2-3B-Instruct  & 3B \\
18 & & & Llama3.2-3B           & 3B \\
\cmidrule{3-5}
19 & & \multirow{3}{*}{Mistral} & Mistral-7B-Instruct-v0\_1 & 7B \\
20 & & & Mistral-7B-Instruct-v0\_2 & 7B \\
21 & & & Mistral-7B-Instruct-v0\_3 & 7B \\
\cmidrule{3-5}
22 & & \multirow{12}{*}{Qwen2.5} & qwen2.5-0.5b-instruct & 0.5B \\
23 & & & qwen2.5-0.5b          & 0.5B \\
24 & & & qwen2.5-1.5b-instruct & 1.5B \\
25 & & & qwen2.5-1.5b          & 1.5B \\
26 & & & qwen2.5-14b-instruct  & 14B \\
27 & & & qwen2.5-14b           & 14B \\
28 & & & qwen2.5-32b-instruct  & 32B \\
29 & & & qwen2.5-32b           & 32B \\
30 & & & qwen2.5-3b-instruct   & 3B \\
31 & & & qwen2.5-3b            & 3B \\
32 & & & qwen2.5-7b-instruct   & 7B \\
33 & & & qwen2.5-7b            & 7B \\
\cmidrule{3-5}
34 & & \multirow{7}{*}{Qwen3} & qwen3-0.6b            & 0.6B \\
35 & & & qwen3-1.7b            & 1.7B \\
36 & & & qwen3-14b             & 14B \\
37 & & & qwen3-30b-a3b         & 30B \\
38 & & & qwen3-32b             & 32B \\
39 & & & qwen3-4b              & 4B \\
40 & & & qwen3-8b              & 8B \\
\cmidrule{3-5}
41 & & QwQ & qwq-32b               & 32B \\
\bottomrule
\end{tabular}
\end{table}

%% file: table/evaluated-benchmarks.tex
\begin{table}[htbp]
\centering
\small
\caption{List of evaluated benchmarks with detailed metadata.}
\label{tab:evaluated-benchmarks}
    \begin{tabular}{clcclcc}
    \toprule
    \textbf{ID} & \textbf{Subject} & \textbf{Name} & \textbf{\# Items} & \textbf{Source} & \textbf{Answer Format} \\
    \midrule
    1 & \multirow{3}{*}{Math} 
    & MMLU-MATH          & 748  & cais/mmlu       & Multiple-choice \\
    2 & & MMLU-Pro-MATH      & 1351 & TIGER-Lab/MMLU-Pro  & Multiple-choice \\
    3 & & MATH500            & 500  & HuggingFaceH4/MATH-500   & Free-response \\
    \midrule
    4 & Chemistry 
    & MMLU-Pro-Chemistry   & 1132 & TIGER-Lab/MMLU-Pro   & Multiple-choice \\
    \midrule
    5 & Physics 
    & MMLU-Pro-Physics     & 1299 & TIGER-Lab/MMLU-Pro   & Multiple-choice \\
    \midrule
    6 & Computer science 
    & MMLU-Pro-CS          & 410  & TIGER-Lab/MMLU-Pro   & Multiple-choice \\
    \bottomrule
    \end{tabular}
\end{table}

%% file: table/distirbution.tex
\begin{table}[htbp]
\centering
\caption{Item coverage of per ability dimension across mathematical benchmarks.}
\label{tab:ability-coverage}
\begin{tabular}{lrrrrrr}
\toprule
\textbf{Dimension} & \multicolumn{2}{c}{\textbf{MMLU-MATH}} & \multicolumn{2}{c}{\textbf{MMLU-Pro-MATH}} & \multicolumn{2}{c}{\textbf{MATH500}} \\
\cmidrule(r){2-3} \cmidrule(r){4-5} \cmidrule(l){6-7}
& \textbf{Count} & \textbf{Ratio} & \textbf{Count} & \textbf{Ratio} & \textbf{Count} & \textbf{Ratio} \\
\midrule
ability\_0  & 61  & 0.0816 & 55  & 0.0407 & 20  & 0.0400 \\
ability\_1  & 340 & 0.4545 & 300 & 0.2221 & 125 & 0.2500 \\
ability\_2  & 77  & 0.1029 & 112 & 0.0829 & 84  & 0.1680 \\
ability\_3  & 183 & 0.2447 & 189 & 0.1399 & 167 & 0.3340 \\
ability\_4  & 23  & 0.0307 & 57  & 0.0422 & 64  & 0.1280 \\
ability\_5  & 41  & 0.0548 & 68  & 0.0503 & 41  & 0.0820 \\
ability\_6  & 36  & 0.0481 & 81  & 0.0600 & 45  & 0.0900 \\
ability\_7  & 37  & 0.0495 & 62  & 0.0459 & 73  & 0.1460 \\
ability\_8  & 30  & 0.0401 & 38  & 0.0281 & 56  & 0.1120 \\
ability\_9  & 49  & 0.0655 & 58  & 0.0429 & 47  & 0.0940 \\
ability\_10 & 11  & 0.0147 & 67  & 0.0496 & 3   & 0.0060 \\
ability\_11 & 30  & 0.0401 & 121 & 0.0896 & 15  & 0.0300 \\
ability\_12 & 16  & 0.0214 & 113 & 0.0836 & 0   & 0.0000 \\
ability\_13 & 15  & 0.0201 & 72  & 0.0533 & 4   & 0.0080 \\
ability\_14 & 38  & 0.0508 & 189 & 0.1399 & 16  & 0.0320 \\
ability\_15 & 3   & 0.0040 & 99  & 0.0733 & 0   & 0.0000 \\
ability\_16 & 18  & 0.0241 & 54  & 0.0400 & 24  & 0.0480 \\
ability\_17 & 8   & 0.0107 & 18  & 0.0133 & 8   & 0.0160 \\
ability\_18 & 12  & 0.0160 & 27  & 0.0200 & 2   & 0.0040 \\
ability\_19 & 28  & 0.0374 & 104 & 0.0770 & 5   & 0.0100 \\
ability\_20 & 60  & 0.0802 & 147 & 0.1088 & 61  & 0.1220 \\
ability\_21 & 3   & 0.0040 & 12  & 0.0089 & 0   & 0.0000 \\
ability\_22 & 2   & 0.0027 & 2   & 0.0015 & 0   & 0.0000 \\
ability\_23 & 19  & 0.0254 & 72  & 0.0533 & 1   & 0.0020 \\
ability\_24 & 40  & 0.0535 & 87  & 0.0644 & 10  & 0.0200 \\
ability\_25 & 40  & 0.0535 & 107 & 0.0792 & 37  & 0.0740 \\
ability\_26 & 42  & 0.0561 & 173 & 0.1281 & 83  & 0.1660 \\
ability\_27 & 442 & 0.5909 & 838 & 0.6203 & 330 & 0.6600 \\
ability\_28 & 103 & 0.1377 & 130 & 0.0962 & 20  & 0.0400 \\
ability\_29 & 77  & 0.1029 & 240 & 0.1776 & 111 & 0.2220 \\
ability\_30 & 1   & 0.0013 & 1   & 0.0007 & 0   & 0.0000 \\
ability\_31 & 0   & 0.0000 & 3   & 0.0022 & 2   & 0.0040 \\
ability\_32 & 0   & 0.0000 & 0   & 0.0000 & 0   & 0.0000 \\
ability\_33 & 0   & 0.0000 & 0   & 0.0000 & 0   & 0.0000 \\
ability\_34 & 1   & 0.0013 & 2   & 0.0015 & 0   & 0.0000 \\
\bottomrule
\end{tabular}
\end{table}

%% file: appendix/prediction.tex
\clearpage

\section{Unseen Item Prediction Details}
\label{app:prediction}

This appendix provides per-model results for unseen item prediction experiments described in Section~\ref{sec:prediction}. For within-benchmark prediction, we perform 10 independent random splits (90\% evaluation set, 10\% prediction set) per model. For cross-benchmark prediction, we use the source benchmark for ability estimation and the target benchmark for prediction. All predictions use fixed item parameters ($I_{\text{diff}} = 0.5$, $I_{\text{disc}} = 1.0$) for unseen items.

\subsection{Within-Benchmark Prediction Performance}
\label{app:within_prediction}

Table~\ref{tab:within_prediction_per_model} reports prediction performance for all 41 models across three mathematical benchmarks. All results substantially exceed trivial baselines (accuracy-based: AUC = 0.500; unidimensional IRT: AUC = 0.500; random prediction: mean AUC $\approx$ 0.49).

\begin{table}[tbp]
\centering
\caption{Within-benchmark unseen item prediction performance per model (AUC $\pm$ std over 10 random splits with different seeds).}
\label{tab:within_prediction_per_model}
\begin{tabular}{lccc}
\toprule
Model & MMLU-MATH & MATH500 & MMLU-Pro-MATH \\
\midrule
0 & 0.8711 $\pm$ 0.0489 & 0.8732 $\pm$ 0.0401 & 0.8522 $\pm$ 0.0303 \\
1 & 0.8672 $\pm$ 0.0356 & 0.8434 $\pm$ 0.0863 & 0.8683 $\pm$ 0.0288 \\
2 & 0.8548 $\pm$ 0.0338 & 0.8049 $\pm$ 0.0681 & 0.8584 $\pm$ 0.0222 \\
3 & 0.8560 $\pm$ 0.0593 & 0.8557 $\pm$ 0.0405 & 0.8573 $\pm$ 0.0398 \\
4 & 0.8583 $\pm$ 0.0607 & 0.8634 $\pm$ 0.0654 & 0.8547 $\pm$ 0.0316 \\
5 & 0.8474 $\pm$ 0.0331 & 0.8586 $\pm$ 0.0552 & 0.8517 $\pm$ 0.0480 \\
6 & 0.8586 $\pm$ 0.0283 & 0.8396 $\pm$ 0.0604 & 0.8474 $\pm$ 0.0485 \\
7 & 0.8761 $\pm$ 0.0265 & 0.8513 $\pm$ 0.0627 & 0.8593 $\pm$ 0.0232 \\
8 & 0.8526 $\pm$ 0.0420 & 0.8552 $\pm$ 0.0534 & 0.8582 $\pm$ 0.0298 \\
9 & 0.8465 $\pm$ 0.0470 & 0.8336 $\pm$ 0.0697 & 0.8595 $\pm$ 0.0288 \\
10 & 0.8599 $\pm$ 0.0554 & 0.8446 $\pm$ 0.0297 & 0.8638 $\pm$ 0.0320 \\
11 & 0.8495 $\pm$ 0.0320 & 0.8459 $\pm$ 0.0413 & 0.8479 $\pm$ 0.0319 \\
12 & 0.8282 $\pm$ 0.0194 & 0.8529 $\pm$ 0.0481 & 0.8437 $\pm$ 0.0133 \\
13 & 0.8248 $\pm$ 0.0553 & 0.8698 $\pm$ 0.0620 & 0.8485 $\pm$ 0.0349 \\
14 & 0.8587 $\pm$ 0.0416 & 0.8816 $\pm$ 0.0483 & 0.8432 $\pm$ 0.0402 \\
15 & 0.8581 $\pm$ 0.0372 & 0.8552 $\pm$ 0.0507 & 0.8551 $\pm$ 0.0180 \\
16 & 0.8526 $\pm$ 0.0392 & 0.8862 $\pm$ 0.0540 & 0.8667 $\pm$ 0.0203 \\
17 & 0.8563 $\pm$ 0.0295 & 0.8774 $\pm$ 0.0322 & 0.8374 $\pm$ 0.0325 \\
18 & 0.8500 $\pm$ 0.0356 & 0.8618 $\pm$ 0.0568 & 0.8625 $\pm$ 0.0385 \\
19 & 0.8273 $\pm$ 0.0334 & 0.8813 $\pm$ 0.0416 & 0.8456 $\pm$ 0.0363 \\
20 & 0.8688 $\pm$ 0.0344 & 0.8449 $\pm$ 0.0335 & 0.8693 $\pm$ 0.0323 \\
21 & 0.8555 $\pm$ 0.0592 & 0.8699 $\pm$ 0.0504 & 0.8557 $\pm$ 0.0488 \\
22 & 0.8612 $\pm$ 0.0534 & 0.8336 $\pm$ 0.0874 & 0.8412 $\pm$ 0.0397 \\
23 & 0.8638 $\pm$ 0.0366 & 0.8516 $\pm$ 0.0545 & 0.8546 $\pm$ 0.0281 \\
24 & 0.8423 $\pm$ 0.0392 & 0.8347 $\pm$ 0.0689 & 0.8398 $\pm$ 0.0254 \\
25 & 0.8378 $\pm$ 0.0486 & 0.8415 $\pm$ 0.0466 & 0.8558 $\pm$ 0.0230 \\
26 & 0.8355 $\pm$ 0.0362 & 0.8317 $\pm$ 0.0513 & 0.8650 $\pm$ 0.0427 \\
27 & 0.8603 $\pm$ 0.0352 & 0.8478 $\pm$ 0.0350 & 0.8586 $\pm$ 0.0329 \\
28 & 0.8429 $\pm$ 0.0544 & 0.8635 $\pm$ 0.0288 & 0.8567 $\pm$ 0.0231 \\
29 & 0.8523 $\pm$ 0.0358 & 0.8268 $\pm$ 0.0786 & 0.8350 $\pm$ 0.0289 \\
30 & 0.8480 $\pm$ 0.0459 & 0.8547 $\pm$ 0.0289 & 0.8588 $\pm$ 0.0346 \\
31 & 0.8401 $\pm$ 0.0356 & 0.8387 $\pm$ 0.0667 & 0.8429 $\pm$ 0.0340 \\
32 & 0.8371 $\pm$ 0.0543 & 0.8941 $\pm$ 0.0290 & 0.8505 $\pm$ 0.0265 \\
33 & 0.8157 $\pm$ 0.0766 & 0.8474 $\pm$ 0.0411 & 0.8343 $\pm$ 0.0342 \\
34 & 0.8548 $\pm$ 0.0604 & 0.8634 $\pm$ 0.0418 & 0.8540 $\pm$ 0.0475 \\
35 & 0.8660 $\pm$ 0.0421 & 0.8782 $\pm$ 0.0305 & 0.8582 $\pm$ 0.0263 \\
36 & 0.8526 $\pm$ 0.0335 & 0.8576 $\pm$ 0.0578 & 0.8462 $\pm$ 0.0277 \\
37 & 0.8703 $\pm$ 0.0518 & 0.8893 $\pm$ 0.0414 & 0.8657 $\pm$ 0.0311 \\
38 & 0.8451 $\pm$ 0.0391 & 0.8799 $\pm$ 0.0467 & 0.8635 $\pm$ 0.0201 \\
39 & 0.8218 $\pm$ 0.0407 & 0.8521 $\pm$ 0.0526 & 0.8587 $\pm$ 0.0439 \\
40 & 0.8576 $\pm$ 0.0375 & 0.8718 $\pm$ 0.0523 & 0.8654 $\pm$ 0.0254 \\
\bottomrule
\end{tabular}
\end{table}

The results demonstrate consistently strong predictive performance across all models and benchmarks, with mean AUC values ranging from 0.8049 to 0.8941. Standard deviations remain modest (typically $<0.06$), indicating stable prediction quality across random item partitions.

\subsection{Cross-Benchmark Prediction Performance}
\label{app:cross_prediction}

Table~\ref{tab:cross_prediction_per_model} reports prediction performance for all 41 models across six ordered benchmark pairs. For each pair, we use the source benchmark for ability estimation and the target benchmark for prediction. All results substantially exceed trivial baselines (accuracy-based: AUC = 0.500; unidimensional IRT: AUC = 0.500; random prediction: mean AUC $\approx$ 0.49).

\begin{table}[tbp]
\centering
\caption{Cross-benchmark unseen item prediction performance per model (AUC $\pm$ std over repeat for 10 times).}
\label{tab:cross_prediction_per_model}
\resizebox{\linewidth}{!}{
\begin{tabular}{lcccccc}
\toprule
Model & MMLU$\rightarrow$MATH500 & MMLU$\rightarrow$MMLU-Pro & MATH500$\rightarrow$MMLU & MATH500$\rightarrow$MMLU-Pro & MMLU-Pro$\rightarrow$MMLU & MMLU-Pro$\rightarrow$MATH500 \\
\midrule
0 & 0.7686 $\pm$ 0.0070 & 0.8345 $\pm$ 0.0035 & 0.8254 $\pm$ 0.0027 & 0.8305 $\pm$ 0.0023 & 0.8448 $\pm$ 0.0057 & 0.7870 $\pm$ 0.0022 \\
1 & 0.7942 $\pm$ 0.0031 & 0.8329 $\pm$ 0.0032 & 0.8132 $\pm$ 0.0017 & 0.8106 $\pm$ 0.0022 & 0.8330 $\pm$ 0.0030 & 0.8287 $\pm$ 0.0045 \\
2 & 0.7766 $\pm$ 0.0033 & 0.8385 $\pm$ 0.0035 & 0.8498 $\pm$ 0.0032 & 0.8203 $\pm$ 0.0017 & 0.8521 $\pm$ 0.0038 & 0.7953 $\pm$ 0.0028 \\
3 & 0.7784 $\pm$ 0.0041 & 0.8425 $\pm$ 0.0046 & 0.8390 $\pm$ 0.0017 & 0.8268 $\pm$ 0.0019 & 0.8636 $\pm$ 0.0046 & 0.7917 $\pm$ 0.0052 \\
4 & 0.8085 $\pm$ 0.0052 & 0.8288 $\pm$ 0.0041 & 0.8299 $\pm$ 0.0019 & 0.8110 $\pm$ 0.0018 & 0.8309 $\pm$ 0.0047 & 0.8201 $\pm$ 0.0028 \\
5 & 0.7774 $\pm$ 0.0036 & 0.8457 $\pm$ 0.0045 & 0.8158 $\pm$ 0.0049 & 0.8172 $\pm$ 0.0020 & 0.8145 $\pm$ 0.0042 & 0.8197 $\pm$ 0.0024 \\
6 & 0.8043 $\pm$ 0.0047 & 0.8336 $\pm$ 0.0044 & 0.8353 $\pm$ 0.0028 & 0.8076 $\pm$ 0.0025 & 0.8488 $\pm$ 0.0048 & 0.8204 $\pm$ 0.0075 \\
7 & 0.8031 $\pm$ 0.0077 & 0.8338 $\pm$ 0.0038 & 0.8099 $\pm$ 0.0018 & 0.8320 $\pm$ 0.0014 & 0.8273 $\pm$ 0.0075 & 0.8346 $\pm$ 0.0040 \\
8 & 0.8107 $\pm$ 0.0058 & 0.8453 $\pm$ 0.0035 & 0.8229 $\pm$ 0.0051 & 0.8310 $\pm$ 0.0018 & 0.8333 $\pm$ 0.0034 & 0.8271 $\pm$ 0.0031 \\
9 & 0.8244 $\pm$ 0.0053 & 0.8338 $\pm$ 0.0023 & 0.8457 $\pm$ 0.0035 & 0.8041 $\pm$ 0.0030 & 0.8464 $\pm$ 0.0015 & 0.8363 $\pm$ 0.0048 \\
10 & 0.8331 $\pm$ 0.0061 & 0.8268 $\pm$ 0.0025 & 0.8240 $\pm$ 0.0027 & 0.8005 $\pm$ 0.0022 & 0.8518 $\pm$ 0.0024 & 0.8420 $\pm$ 0.0015 \\
11 & 0.8056 $\pm$ 0.0036 & 0.8421 $\pm$ 0.0020 & 0.8305 $\pm$ 0.0017 & 0.8242 $\pm$ 0.0023 & 0.8362 $\pm$ 0.0044 & 0.8310 $\pm$ 0.0034 \\
12 & 0.7813 $\pm$ 0.0057 & 0.8362 $\pm$ 0.0051 & 0.8208 $\pm$ 0.0028 & 0.8187 $\pm$ 0.0016 & 0.8393 $\pm$ 0.0049 & 0.7841 $\pm$ 0.0043 \\
13 & 0.7821 $\pm$ 0.0042 & 0.8438 $\pm$ 0.0031 & 0.8361 $\pm$ 0.0038 & 0.8192 $\pm$ 0.0021 & 0.8476 $\pm$ 0.0050 & 0.8145 $\pm$ 0.0038 \\
14 & 0.7810 $\pm$ 0.0061 & 0.8286 $\pm$ 0.0017 & 0.8297 $\pm$ 0.0029 & 0.8161 $\pm$ 0.0022 & 0.8459 $\pm$ 0.0040 & 0.8027 $\pm$ 0.0021 \\
15 & 0.8053 $\pm$ 0.0045 & 0.8556 $\pm$ 0.0024 & 0.8454 $\pm$ 0.0032 & 0.8297 $\pm$ 0.0024 & 0.8413 $\pm$ 0.0045 & 0.8338 $\pm$ 0.0074 \\
16 & 0.8003 $\pm$ 0.0055 & 0.8275 $\pm$ 0.0026 & 0.8347 $\pm$ 0.0033 & 0.8031 $\pm$ 0.0022 & 0.8643 $\pm$ 0.0025 & 0.8333 $\pm$ 0.0023 \\
17 & 0.7788 $\pm$ 0.0051 & 0.8513 $\pm$ 0.0018 & 0.8094 $\pm$ 0.0024 & 0.8274 $\pm$ 0.0017 & 0.8105 $\pm$ 0.0084 & 0.8071 $\pm$ 0.0040 \\
18 & 0.8091 $\pm$ 0.0043 & 0.8410 $\pm$ 0.0029 & 0.8342 $\pm$ 0.0024 & 0.8162 $\pm$ 0.0023 & 0.8351 $\pm$ 0.0061 & 0.8336 $\pm$ 0.0041 \\
19 & 0.7826 $\pm$ 0.0084 & 0.8438 $\pm$ 0.0029 & 0.8174 $\pm$ 0.0043 & 0.8128 $\pm$ 0.0034 & 0.8459 $\pm$ 0.0021 & 0.7979 $\pm$ 0.0051 \\
20 & 0.7935 $\pm$ 0.0037 & 0.8292 $\pm$ 0.0023 & 0.8198 $\pm$ 0.0020 & 0.8203 $\pm$ 0.0021 & 0.8344 $\pm$ 0.0052 & 0.8146 $\pm$ 0.0032 \\
21 & 0.7901 $\pm$ 0.0022 & 0.8296 $\pm$ 0.0023 & 0.8461 $\pm$ 0.0020 & 0.8093 $\pm$ 0.0018 & 0.8408 $\pm$ 0.0029 & 0.8085 $\pm$ 0.0026 \\
22 & 0.8144 $\pm$ 0.0050 & 0.8319 $\pm$ 0.0033 & 0.8453 $\pm$ 0.0018 & 0.8202 $\pm$ 0.0012 & 0.8542 $\pm$ 0.0075 & 0.8350 $\pm$ 0.0025 \\
23 & 0.8009 $\pm$ 0.0068 & 0.8330 $\pm$ 0.0036 & 0.8438 $\pm$ 0.0022 & 0.8216 $\pm$ 0.0019 & 0.8474 $\pm$ 0.0029 & 0.8339 $\pm$ 0.0023 \\
24 & 0.7912 $\pm$ 0.0061 & 0.8350 $\pm$ 0.0038 & 0.8169 $\pm$ 0.0038 & 0.8130 $\pm$ 0.0032 & 0.8179 $\pm$ 0.0054 & 0.8259 $\pm$ 0.0032 \\
25 & 0.7684 $\pm$ 0.0043 & 0.8352 $\pm$ 0.0030 & 0.8218 $\pm$ 0.0035 & 0.8144 $\pm$ 0.0024 & 0.8479 $\pm$ 0.0026 & 0.7867 $\pm$ 0.0031 \\
26 & 0.7891 $\pm$ 0.0031 & 0.8349 $\pm$ 0.0033 & 0.8253 $\pm$ 0.0034 & 0.8121 $\pm$ 0.0023 & 0.8374 $\pm$ 0.0049 & 0.8025 $\pm$ 0.0030 \\
27 & 0.7930 $\pm$ 0.0047 & 0.8331 $\pm$ 0.0036 & 0.8231 $\pm$ 0.0041 & 0.8111 $\pm$ 0.0028 & 0.8234 $\pm$ 0.0065 & 0.8132 $\pm$ 0.0051 \\
28 & 0.8026 $\pm$ 0.0062 & 0.8062 $\pm$ 0.0047 & 0.8139 $\pm$ 0.0025 & 0.8046 $\pm$ 0.0019 & 0.8330 $\pm$ 0.0059 & 0.8006 $\pm$ 0.0030 \\
29 & 0.7743 $\pm$ 0.0061 & 0.8439 $\pm$ 0.0035 & 0.8204 $\pm$ 0.0027 & 0.8236 $\pm$ 0.0017 & 0.8353 $\pm$ 0.0054 & 0.8045 $\pm$ 0.0031 \\
30 & 0.7927 $\pm$ 0.0054 & 0.8387 $\pm$ 0.0037 & 0.8349 $\pm$ 0.0020 & 0.8299 $\pm$ 0.0019 & 0.8441 $\pm$ 0.0046 & 0.8014 $\pm$ 0.0032 \\
31 & 0.8067 $\pm$ 0.0048 & 0.8355 $\pm$ 0.0029 & 0.8484 $\pm$ 0.0018 & 0.8123 $\pm$ 0.0022 & 0.8461 $\pm$ 0.0051 & 0.8331 $\pm$ 0.0068 \\
32 & 0.8078 $\pm$ 0.0061 & 0.8262 $\pm$ 0.0021 & 0.8114 $\pm$ 0.0022 & 0.8261 $\pm$ 0.0024 & 0.8034 $\pm$ 0.0081 & 0.8292 $\pm$ 0.0031 \\
33 & 0.7914 $\pm$ 0.0052 & 0.8373 $\pm$ 0.0046 & 0.8156 $\pm$ 0.0023 & 0.8298 $\pm$ 0.0023 & 0.8270 $\pm$ 0.0050 & 0.8077 $\pm$ 0.0025 \\
34 & 0.7742 $\pm$ 0.0056 & 0.8273 $\pm$ 0.0032 & 0.8274 $\pm$ 0.0020 & 0.8145 $\pm$ 0.0022 & 0.8442 $\pm$ 0.0040 & 0.7980 $\pm$ 0.0027 \\
35 & 0.7920 $\pm$ 0.0051 & 0.8433 $\pm$ 0.0025 & 0.8149 $\pm$ 0.0010 & 0.8297 $\pm$ 0.0013 & 0.8113 $\pm$ 0.0059 & 0.8069 $\pm$ 0.0039 \\
36 & 0.7718 $\pm$ 0.0059 & 0.8349 $\pm$ 0.0038 & 0.8175 $\pm$ 0.0017 & 0.8051 $\pm$ 0.0014 & 0.8222 $\pm$ 0.0043 & 0.7898 $\pm$ 0.0062 \\
37 & 0.7820 $\pm$ 0.0056 & 0.8402 $\pm$ 0.0044 & 0.8188 $\pm$ 0.0034 & 0.8103 $\pm$ 0.0011 & 0.8378 $\pm$ 0.0039 & 0.7968 $\pm$ 0.0037 \\
38 & 0.7688 $\pm$ 0.0077 & 0.8253 $\pm$ 0.0030 & 0.8042 $\pm$ 0.0050 & 0.7895 $\pm$ 0.0021 & 0.8273 $\pm$ 0.0043 & 0.7958 $\pm$ 0.0031 \\
39 & 0.8058 $\pm$ 0.0035 & 0.8426 $\pm$ 0.0028 & 0.8179 $\pm$ 0.0041 & 0.8225 $\pm$ 0.0028 & 0.8344 $\pm$ 0.0056 & 0.8267 $\pm$ 0.0044 \\
40 & 0.7864 $\pm$ 0.0067 & 0.8444 $\pm$ 0.0031 & 0.7991 $\pm$ 0.0020 & 0.8267 $\pm$ 0.0029 & 0.8142 $\pm$ 0.0087 & 0.8070 $\pm$ 0.0015 \\
\bottomrule
\end{tabular}
}
\end{table}

The results demonstrate consistent cross-benchmark generalization with mean AUC values ranging from 0.7684 to 0.8556. Performance remains stable across all model-benchmark pairs, with standard deviations typically $<0.008$, confirming the framework's robustness in transferring ability estimates across domains.